\documentclass{article} % For LaTeX2e
\usepackage{iclr2023_conference,times}
\usepackage{booktabs}
\usepackage{caption}
\usepackage{subcaption}
\usepackage{graphicx}
\usepackage{subfiles}
\usepackage{wrapfig,lipsum,booktabs}
\usepackage{xcolor}

\usepackage{amsmath,amsfonts,bm}

\def\eqref#1{equation~\ref{#1}}
\def\1{\bm{1}}

\DeclareMathAlphabet{\mathsfit}{\encodingdefault}{\sfdefault}{m}{sl}
\SetMathAlphabet{\mathsfit}{bold}{\encodingdefault}{\sfdefault}{bx}{n}

\usepackage{hyperref}
\usepackage{url}

\title{Prompting GPT-3 To Be Reliable}

\author{Chenglei Si$^1$\thanks{Work done during internship at Microsoft.} , ~Zhe Gan$^2$, ~Zhengyuan Yang$^2$, ~Shuohang Wang$^2$ \\ 
\textbf{~Jianfeng Wang$^2$, Jordan Boyd-Graber$^{1}$, Lijuan Wang$^2$} \\
    $^1$ University of Maryland \hspace{0.4cm}
    $^2$ Microsoft\\%
    \hspace{0.15cm} \texttt{clsi@umd.edu} \hspace{0.2cm} \texttt{pkuganzhe@gmail.com} 
    \hspace{0.2cm} \texttt{lijuanw@microsoft.com }
}

\iclrfinalcopy % Uncomment for camera-ready version, but NOT for submission.
\begin{document}

\maketitle

\begin{abstract}
Large language models (LLMs) show impressive abilities via few-shot prompting. 
Commercialized APIs such as OpenAI GPT-3 further increase their use in real-world language applications.
%
% However, existing research focus on models' accuracy on standard benchmarks and largely ignore their \emph{reliability}, which is crucial for avoiding catastrophic real-world harms.  
However, the crucial problem of how to improve the \emph{reliability} of GPT-3 is still under-explored.
While reliability is a broad and vaguely defined term, we decompose reliability into four main facets that correspond to the existing framework of ML safety and are well-recognized to be important: generalizability, social biases, calibration, and factuality.
Our core contribution is to establish simple and effective prompts that
improve GPT-3's reliability as it:
1) generalizes out-of-distribution, 
2) balances demographic distribution and uses natural language instructions to reduce social biases, 
3) calibrates output probabilities, and
4) updates the LLM's factual knowledge and reasoning chains.
With appropriate prompts, GPT-3 is more reliable than smaller-scale supervised models on all these facets.
We release all processed datasets, evaluation scripts, and model predictions.\footnote{\url{https://github.com/NoviScl/GPT3-Reliability}}
Our systematic empirical study not only sheds new insights on the reliability of prompting LLMs, but more importantly, our prompting strategies can help practitioners more reliably use LLMs like GPT-3.
\end{abstract}

%\vspace{-2mm}
\section{Introduction}
%\vspace{-2mm}

% Introduce LLMs, motivate why reliability is important, mention related works on reliability testing 
NLP is dominated by large language models (LLMs) --- pretrained on large, unlabeled text data --- that are then used for downstream tasks~\citep{BERT,GPT3}.
Scaling the model and data size often brings gains on downstream tasks~\citep{Scaling,BIG-Bench}, allowing what some call emergent abilities~\citep{Wei2022EmergentAO}. 
These emergent behaviors are accomplished through prompting---a crafted, natural language text to shape predictions or offer relevant information without expensive supervised data.
Among all the existing LLMs, GPT-3~\citep{GPT3} is particularly popular due to its flexibility and ease of use from the OpenAI API~\footnote{By default, we use the \textsc{code-davinci-002} model (also known as Codex or GPT 3.5) in our experiments unless otherwise specified, because our preliminary results show that this is the most accurate model on most NLP datasets we tried. 
% The official documentation did not specify the exact size of \textsc{code-davinci-002}, but we expect it to be at least the billion scale and much larger than all supervised baselines used in this paper. 
% We specify the number of parameters for all other models being used in the experiment result sections.
}.

Existing empirical studies investigate GPT-3 on specific tasks such as mathematical reasoning~\citep{Hendrycks2021MeasuringMP}, multi-hop reasoning~\citep{Wei2022ChainOT,Kojima2022LargeLM}, and code generation~\citep{Codex}.
However, rising numbers on these evaluations do not ensure LLM \emph{reliability}.
For example, LLMs (including GPT-3) produce biased~\citep{Lucy2021GenderAR} 
% or privacy-leaking~\citep{Carlini2021ExtractingTD} 
generations, false statements~\citep{Lin2022TruthfulQAMH}, 
and outdated information~\citep{Chen2021ADF,Kasai2022RealTimeQW}.
Deploying such models in the real world could result in catastrophic harm.
% a model in the real world, especially in high-stakes settings, could result in catastrophic harm.
% ~\citep{Amodei2016ConcretePI,Weidinger2021EthicalAS}. 
%
% Therefore, it is crucial to systematically assess and improve model reliability; and this work focuses on the widely used GPT-3 models.\footnote{By default, we use the \textsc{code-davinci-002} model in our experiments unless otherwise specified. This choice is because our preliminary results show that this is the best-performing model on most NLP datasets we have tried, and more closely represents the state-of-the-art few-shot model.}

\begin{figure}[t]
\centering
\includegraphics[width=1.0\textwidth]{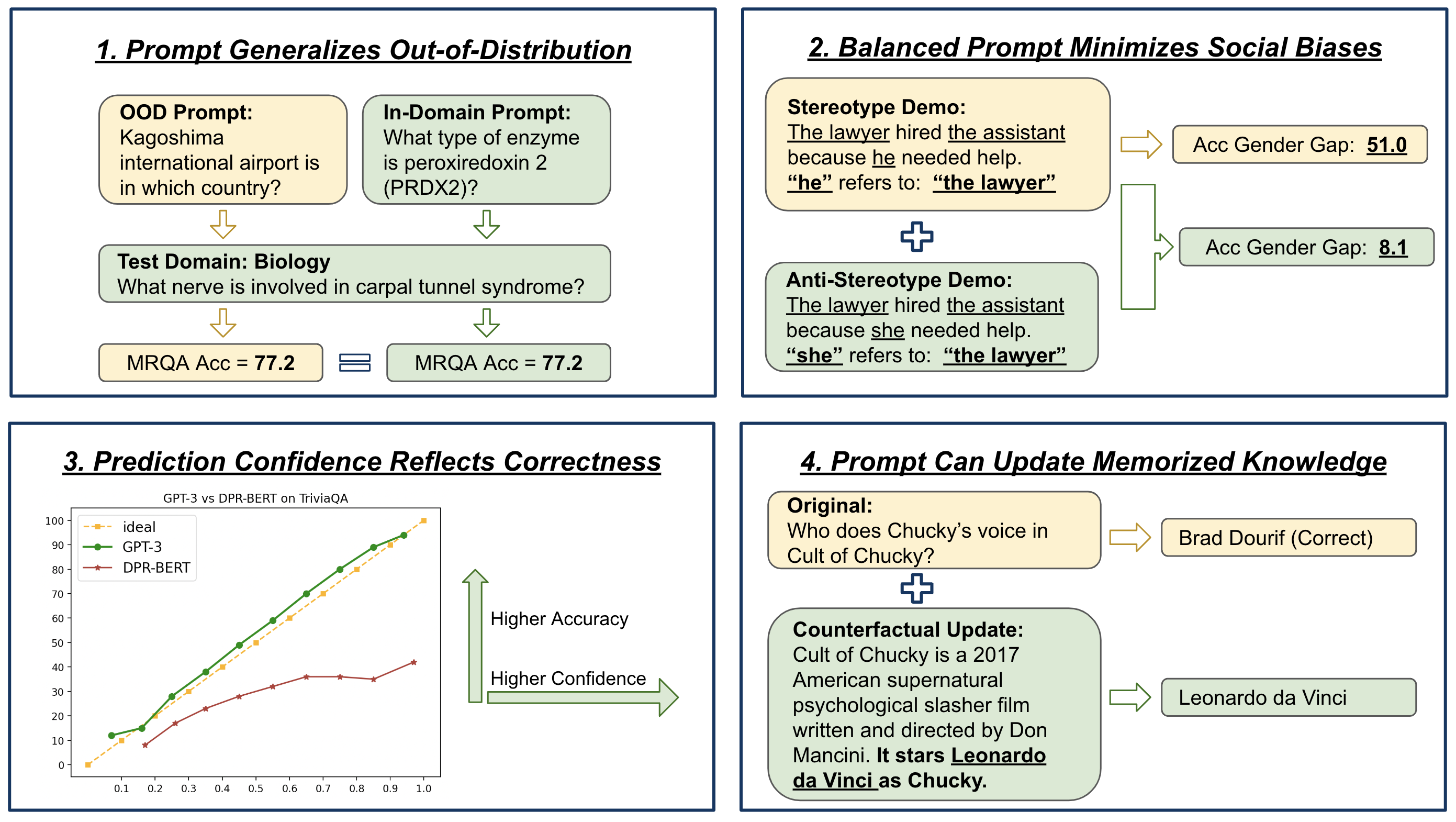}
\caption{Four main reliability factors we examined and the core findings.
% \shuohang{For figure 1: could you replace OOD and In-Domain with a specific domain, such as biology. Is MRQA an overall performance or only for biology? If it's an overall performance, maybe we can change "test domain: biology" to "multi-domain test set"? Figure 2 is not very clear. Maybe add some short description in the caption. Figure3, maybe explain what's confidence? Is it only based on LM probability? Figure 4 is  not very clear. What is  counterfactual update? After adding it, the answer is wrong? Maybe also add some explanation in caption.}
}
\label{fig:Fig1}
% \vspace{-1mm}
\end{figure}

In the context of prompting LLMs, several previous works have explored their reliability. For example, in the release reports of GPT-3~\citep{GPT3}, OPT~\citep{Zhang2022OPTOP}, Gopher~\citep{Rae2021ScalingLM} and PaLM~\citep{Chowdhery2022PaLMSL}, there are dedicated experiments evaluating these LLMs' representational bias and toxicity. Another line of work has  evaluated calibration~\citep{Lin2022TeachingMT,Kadavath2022LanguageM} of prompting-based LLMs on math questions or multiple-choice questions. 
We differ from these prior works in two key aspects: ($i$) We perform a more comprehensive study of four core facets of reliability, serving as a meta-analysis. ($ii$) We focus particularly on finding prompting strategies that are effective under these reliability facets, rather than just evaluating intrinsic model characteristics (Figure~\ref{fig:Fig1}). 

% Previous work have considered the  reliability of LLMs. 
% for example, through automatic stress tests~\citep{Ribeiro2020BeyondAB}, adversarial attacks~\citep{Wallace2019UniversalAT}, and human red-teaming~\citep{Ganguli2022RedTL}.~\footnote{We provide a more extensive discussion of related works in Appendix~\ref{sec:related_work}.}
% %
% They focus on specific aspects and their evaluation methods may not feasible on the much larger GPT-3 model. 
%

% To paint a more comprehensive picture, we decompose reliability into four  \textbf{facets}: 1) generalizability to distribution shifts; 2) social biases and fairness; 3) uncertainty calibration; and 4) factual correctness. 
% We motivate and elaborate on each of these four aspects in Section~\ref{sec:framework}. 
% We highlight that our proposed evaluation framework is just a starting point toward comprehensive reliability testing: these factors should be considered necessary but not sufficient factors for ensuring reliability. In fact, we hope our work can motivate more future work to further substantiate our reliability testing framework with more relevant factors. 
%
% Throughout our evaluation, we focus on the few-shot in-context prompting paradigm because this is how GPT-3 is typically used in downstream tasks. 
%
% In our reliability testing framework, 
% we benchmark GPT-3's reliability and engineer prompts to induce reliable behaviors in some challenging settings. 
%

Our reliability testing framework takes inspiration from the survey of unsolved problems in ML safety~\citep{Hendrycks2021UnsolvedPI}: withstanding hazards (generalizability), identifying hazards (calibration), steering ML systems and reducing deployment hazards (reducing social biases and improving factuality). These facets also aim to address the risks of ML systems identified in existing conceptual frameworks~\citep{Tan2022TheRO,Tan2021ReliabilityTF}.
\textbf{We have a more extensive discussion of related works in Appendix Section~\ref{sec:related_work}.}

As summarized in Figure~\ref{fig:Fig1}, our simple prompting strategies beat smaller-scale supervised models on all reliability metrics we consider: 
1) prompting with randomly sampled examples from the source domain allows GPT-3 to generalize robustly on unseen domains and challenge examples;
2) examples sampled from a balanced demographic distribution and natural language intervention reduce social biases; 
3) language model probabilities are calibrated to reflect accuracy; and 
4) appending up-to-date knowledge can supplant GPT-3's memorized knowledge or reasoning chains. 
%
% In particular, while we find that GPT-3 can be robust to distribution shifts between test data and the prompt demonstrations, it is surprisingly good at capturing patterns in the prompt. For example, a prompt with an imbalanced gender group representation or biased answers can easily lead GPT-3 to produce highly biased predictions; and a prompt with new knowledge updates can lead the model to produce updated answers and forget about its original memorized answer. 
% In the rest of the paper, we detail these experiments in Sections \ref{sec:robustness}-\ref{sec:knowledge_updating} along with a summary in Section \ref{sec:conclusion} containing concrete takeaways for practitioners and outlines for future work. 
% \jbgcomment{Remove the forward points from this sentence and distribute through the introduction where they are most applicable}

% \zhe{we can use vspace to get some additional space throughout the paper.}

%\vspace{-2mm}
\section{Facet 1: Generalizability}\label{sec:robustness}
%\vspace{-2mm}
 % In this section, we introduce our evaluation setup and experiment results for measuring GPT-3's robustness to distribution shifts. 

 % \chenglei{TODO: Add a Figure to show prompt and examples from the three sets of benchmarks.}

 % \subsection{Evaluation Setups}

LLMs are often criticized for missing the forest for the trees.  They overfit training data from a particular domain (domain shift), are not robust to minor changes in a text (perturbations), or use shortcuts to make predictions (spurious correlations).  
These pathologies make models unreliable since these distribution shifts happen all the time in real-world data and could incur significant performance drops. 
In this section, we study whether GPT-3 can stay robust when the test data come from different distributions than the demo examples in the prompt, and how their generalization compares to supervised models. 
% For each of these, researchers have created datasets that target modern models' weaknesses which we adopt for evaluation. 

\paragraph{Experiment Setup}
We study all three types of distribution shifts mentioned above. 
For each of them, researchers have created datasets that target modern language models' weaknesses which we adopt for evaluation. 
For \textbf{domain shift}, MRQA~\citep{Fisch2019MRQA2S} trains on six machine reading datasets from the source domain and tests on six different target domains; for \textbf{perturbations}, AdvGLUE~\citep{Wang2021AdversarialGA} craft adversarial versions of  GLUE~\citep{Wang2018GLUEAM} based on automatic adversarial perturbations and human filtering, and Contrast Sets~\citep{Gardner2020EvaluatingML} are expert-authored minimal edits that change the label; for \textbf{spurious correlation}, HANS~\citep{HANS} and PAWS~\citep{PAWS} are challenge sets designed for models trained on MNLI~ and QQP where the lexical overlap feature in the training data does not hold during testing. 
For each of these settings, we evaluate a simple prompting strategy by 
sampling examples from the source domains (for MRQA, we use a fixed prompt consisting of eight randomly sampled examples from the source domain on all target datasets; 
for perturbations and spurious correlation, we randomly sample 16 demos from the original clean training data from GLUE, MNLI, and QQP respectively).  In addition, for domain shift, we also consider a prompt where we sample eight examples from the training set of each target domain to ablate the impact of the distribution of the demo examples.

\begin{table*}[t]
\small
\begin{center}
\setlength{\tabcolsep}{1mm}{
\begin{tabular}{ l  c c c | c c c | c c c }
\toprule
& \multicolumn{3}{c}{MRQA} & \multicolumn{3}{c}{AdvGLUE} & \multicolumn{3}{c}{Contrast Set} \\
\cmidrule(lr){2-4} \cmidrule(lr){5-7} \cmidrule(lr){8-10}
& Source$_{\uparrow}$ & Target$_{\uparrow}$ & Gap$_{\downarrow}$ & Original$_{\uparrow}$ & Perturbed$_{\uparrow}$ & Gap$_{\downarrow}$ & Original$_{\uparrow}$ & Perturbed$_{\uparrow}$ & Gap$_{\downarrow}$ \\
\midrule
RoBERTa & 81.6 & 62.1 & 19.5 & 91.7 & 51.7 & 40.0 & 86.1 & 71.1 & 15.0 \\
GPT-3 & 79.8 & 77.2 (S) / 77.2 (T) & \textbf{\underline{2.6}} & 84.2 & 69.3 & \textbf{\underline{14.9}} & 85.5 & 80.0 & \textbf{\underline{5.5}} \\
 \bottomrule
\end{tabular}}
 \caption{For GPT-3 on MRQA target domain test sets, we report results for using both demos from the source (S) and target (T) domains, which surprisingly achieve the same F1 on the target domains (77.2). For AdvGLUE and Contrast Set, we use accuracy as the metric and we use demos from the clean data as the prompt. In all cases, GPT-3 few-shot prompting incurs much smaller performance gaps on the OOD or challenge test sets than the supervised RoBERTa (123M)  baseline. We include the comparison with many other supervised baselines in Appendix~\ref{appendix:generalization_baselines} and GPT-3 exhibits better generalization than all of them.}
 \label{tab:robustness}
 \vspace{-1.5mm}
\end{center}
\end{table*}

% \jbgcomment{I think the biggest weakness of the writing of the paper at the moment is that is reads like a list of results.  There needs to be a clear thread throughout with the clear message of the paper that's repeated again and again.  I'm not sure what that message is yet: maybe something like ``Little tweaks to prompts can address lots of problems people see in LLMs: here's a way that anyone can address these issues.''  This should be the framing of every section: A says this is a problem measured on datasets X, Y, Z, here's a prompt that fixes the problem.}
% \chenglei{I agree, although I think this is worth more discussion which I don't have time now. Let's have this discussion post-submission.}

\begin{wraptable}{R}{5.8cm}
\vspace{-4mm}
\begin{tabular}{ l c c c }
\toprule
    & BERT  & RoBERTa  & GPT-3  \\
    & \small{(340M)} & \small{(354M)} & \\
    \midrule
\multicolumn{4}{c}{\textit{MNLI} $\rightarrow$ \textit{HANS}} \\
\midrule
MNLI$_{\uparrow}$ & 86.2 & 89.1 & 77.6 \\
HANS$_{\uparrow}$ & 71.4 & 77.1 & 75.3 \\
Gap$_{\downarrow}$ & 14.8 & 12.0 & \textbf{\underline{2.3}} \\
\midrule
\multicolumn{4}{c}{\textit{QQP} $\rightarrow$ \textit{PAWS}} \\
\midrule
QQP$_{\uparrow}$ & 91.3 & 89.0 & 83.5 \\
PAWS$_{\uparrow}$ & 40.1 & 39.5 & 73.7 \\
Gap$_{\downarrow}$ & 51.2 & 49.5 & \textbf{\underline{9.8}} \\
 \bottomrule
\end{tabular}
\caption{When using demos sampled from MNLI and QQP, GPT-3 few-shot prompting achieves much better generalization than smaller supervised models (BERT and RoBERTa) on the OOD test sets HANS and PAWS.}\label{tab:spurious_small}
\vspace{-3mm}
\end{wraptable}

% \paragraph{Spurious Correlation} 
% Spurious correlations are shortcuts models exploit which may not hold true during testing. 
% %
% We adopt two widely used evaluation setups:
% MNLI-HANS and QQP-PAWS.
% %
% HANS~\citep{HANS} exploits high lexical overlap between sentence pairs often corresponds to entailment; models trained on MNLI exploit this spurious feature and thus fail on HANS examples  constructed from expert-designed templates.
% %
% Similarly, lexical overlap is a common spurious feature in QQP (paraphrase detection) where high lexical overlap mostly leads to paraphrased questions.
% %
% PAWS~\citep{PAWS} is created via automatic word scrambling and human filtering to construct counter-examples where high lexical overlap examples can also correspond to non-paraphrased questions. 
% %
% For GPT-3 prompting,  we randomly sample 16 demo examples (per class) from the MNLI and QQP training sets as the prompt. 
 % In this case, the demo examples simulate the distribution of the bias features in the original training data since they are randomly sampled with uniform distribution. 
 % We perform both random sampling and targeted sampling for ablation analysis as detailed below. 

% \jbgcomment{Now that I've gone through the three datasets, I think this could be simplifed even further.  Here's a rough draft.

% \chenglei{TODO: if need more space, follow Jordan's advice here.}

% \subsection{Results}
\paragraph{Results}
Table~\ref{tab:robustness} and Table~\ref{tab:spurious_small} compare supervised  RoBERTa~\citep{Liu2019RoBERTaAR} and BERT~\citep{Devlin2019BERTPO} models trained on the source domain datasets or the clean training data with GPT-3 that uses examples sampled from the same training data as in the supervised models.~\footnote{For Contrast Set, we show results on the BoolQ subset in the main paper and present results for other subsets in Table~\ref{tab:contrast_sets_full}, which gives the same conclusion that GPT-3 is more robust than supervised models.} GPT-3 achieves higher accuracy on the OOD tests even when it is slightly worse on the in-domain test sets than the supervised baselines, leading to smaller generalization gaps. This shows that prompting GPT-3 can be more robust than supervised finetuning of smaller-scale language models. Surprisingly, we compare using demo examples sampled from the source domains versus target domains on MRQA, and both prompting methods give the same OOD generalization results, indicating that GPT-3 prompts can directly generalize to OOD test sets where the test examples are from a different distribution than the prompt demo distribution, possibly because the role of demonstration examples is more in specifying the task rather than informing the input distribution~\citep{Min2022RethinkingTR}. 

\paragraph{Takeaway} 
 ($i$) Few-shot prompting of GPT-3 is more robust than supervised models such as finetuned BERT and RoBERTa, under all three settings (domain shift, perturbations, spurious correlation). ($ii$) Using randomly sampled demos from the source datasets is a simple but strong baseline, in fact, it performs the same as using demos sampled from the target distributions.

\section{Facet 2: Social Bias and Fairness}
%\vspace{-2mm}

Apart from high performance on in-domain and OOD datasets, the second key facet of reliability is that we expect models to be fair to different demographic groups. 
Biased models cause severe harm when deployed in real-world applications, especially to the minority groups being discriminated against~\citep{Cao2022OnTI}.
In this section,  we examine whether GPT-3 produces biased predictions in two downstream tasks - coreference resolution and question answering. 
% which is of greater practical importance than measuring the intrinsic biases in the internal model representations~\citep{Cao2022OnTI}. 
% TODO: add this paper to related work
% We use two benchmarks grounded on actual downstream tasks: WinoBias and BBQ. We also explore prompting strategies to mitigate biased predictions. 

\subsection{The Case of Gender Bias: WinoBias}

\paragraph{Dataset} 
We start with the WinoBias dataset~\citep{Zhao2018GenderBI} which uses templates to check whether models are more likely to assign gender pronouns to stereotypical occupations. WinoBias has two types of examples: Type I are ambiguous, challenging examples that require world knowledge; Type II can be resolved using only syntactic information. For each type, examples either confirm (pro-bias) or challenge (anti-bias) societal biases.
Ideally, coreference accuracy should be similar on the pro-bias and anti-bias subsets (small gaps).  

\paragraph{Prompt Design} For ease of evaluation, we re-format the WinoBias dataset into a question-answering format where we provide the original sentence and then add a question \textit{``What does the pronoun refer to in the above sentence?''} (\textit{``the pronoun''} is replaced with the actual pronoun in the sentence) and we use the answer exact match as the evaluation metric. We randomly sample examples from the training set as the prompt and then evaluate on the Pro and Anti test sets. 
% We discuss \emph{which} examples are used in the prompt next.
% We explore several different ways to configure the prompt in order to analyse the impact of the prompt, as detailed below. 

% \paragraph{Baselines} We include results of a neural coreference resolution model (E2E) taken from \cite{Zhao2018GenderBI}. Note that these results are not comparable with GPT-3 results because E2E results are measured with the original coreference resolution task format while we cast the data into a free-form QA format for evaluation. The purpose of showing this baseline is simply to illustrate that previous models do exhibit significant gender biases as measured by WinoBias.  

\begin{table}[t]
\small
\begin{center}
\setlength{\tabcolsep}{1.8mm}{
\begin{tabular}{ l | c c c | c c c}
\toprule
  Prompt & Type I Pro$_{\uparrow}$ &  Type I Anti$_{\uparrow}$ & Gap$_{|\downarrow|}$ & Type II Pro$_{\uparrow}$ & Type II Anti$_{\uparrow}$ & Gap$_{|\downarrow|}$ \\
  \midrule
  \multicolumn{7}{c}{Supervised Baseline} \\
  \midrule
    E2E~\citep{Lee2017EndtoendNC}  & 74.9 & 47.4 & 27.2 & 88.6 & 77.3 & 11.3 \\
    \midrule
  \multicolumn{7}{c}{GPT-3 Few-Shot: \textit{Bias Distribution in the Prompt (16 shots)}} \\
\midrule
Balanced & 89.2 & 81.1 & 8.1 & 99.2 & 95.5 & 3.7 \\
Type I - Pro & 93.4 & 42.4 & 51.0 & 91.1 & 78.9 & 12.2 \\
Type II - Pro & 87.6 & 59.5 & 28.1 & 100.0 & 98.7 & 1.3 \\
Type I - Anti & 50.8 & 80.8 & -30.0 & 57.4 & 51.1 & 6.3 \\
Type II - Anti & 85.5 & 68.2 & 17.3 & 100.0 & 99.5 & 0.5 \\
\midrule
\multicolumn{7}{c}{GPT-3 Few-Shot: \textit{Prompt Ordering (16 shots, Balanced)}} \\
\midrule
Randomly Shuffled & 89.2 & 81.1 & 8.1 & 99.2 & 95.5 & 3.7 \\
Pro in the end & 89.5 & 76.3 & 13.2 & 93.7 & 81.8 & 11.9 \\
Anti in the end & 94.2 & 73.2 & 21.0 & 95.5 & 87.1 & 8.4 \\
% \midrule
% \multicolumn{7}{c}{\textit{Impact of Number of Shots}} \\
% \midrule
% Balanced, 16-shots & 89.2 & 81.1 & 8.1 & 99.2 & 95.5 & 3.7\\
% Balanced, 32-shots & 90.8 & 77.4 & 13.4 & 99.7 & 98.9 & 0.8 \\
% Type I - Pro, 16-shots & 93.4 & 42.4 & 51.0 & 91.1 & 78.9 & 12.2 \\
% Type I - Pro, 32-shots & 92.3 & 37.1 & 55.2 & 94.5 & 84.7 & 9.8 \\
 \bottomrule
\end{tabular}}
 \caption{GPT-3 results on WinoBias. The bias gap between Pro-Bias subsets and Anti-Bias subsets indicates the extent of gender biases exhibited by the model (smaller-scale gap is better).
 Results of the baseline ECE model~\citep{Lee2017EndtoendNC} are taken from the WinoBias paper~\citep{Zhao2018GenderBI}. 
 Prompt with balanced pro-bias and anti-bias answers best shrinks the bias gap, and randomly shuffling the demos is better than putting one group at the end.
 } 
 \label{tab:wino_bias}
 \vspace{-2mm}
\end{center}
\end{table}

\paragraph{Which Examples Should be in the Prompt} We compare: 1) sampling four demo examples from each of the Type I-Pro, Type I-Anti, Type II-Pro, and Type II-Anti subsets (Balanced), which results in a total of 16 demos; 2) sampling 16 demo examples from a single subset. The balanced prompt induces the least biased predictions (Table~\ref{tab:wino_bias}, second block). In particular, if we only keep Pro-Bias examples, the model will favor Pro-Bias predictions (especially on Type I test examples because they are more ambiguous while Type II examples have clear syntax cues). % , and similar effect can be observed for the case of using only Anti-Bias examples in the prompt. 

\paragraph{How Should Examples be Ordered}
We compare: 1) randomly shuffling the demo examples; and 2) putting all Pro-Bias or Anti-Bias examples at the end of the prompt. 
% 3) putting all Anti-Bias examples at the end of the prompt. 
Random shuffling reduces bias gaps most (Table~\ref{tab:wino_bias},
third block).  
Interestingly, putting either Pro-Bias or Anti-Bias
examples at the end increases bias gaps. 

\begin{table}[t]
\small
\begin{center}
\setlength{\tabcolsep}{2mm}{
\begin{tabular}{ l  c c c  c }
\toprule
  Prompt & Ambig Acc$_{\uparrow}$ & DisAmbig Acc$_{\uparrow}$ & Ambig Bias Score$_{|\downarrow|}$ & DisAmbig Bias Score$_{|\downarrow|}$ \\
  \midrule
  \multicolumn{5}{c}{\textit{Supervised Baselines}} \\
\midrule
RoBERTa-Base \small{(123M)} & 61.2 & 52.7 & 4.9 & 4.7 \\
RoBERTa-Large \small{(354M)} & 49.4 & 87.3 & 10.4 & 1.2 \\
DeBERTa-Base \small{(184M)} & 47.6 & 90.4 & 12.8 & 2.9 \\
DeBERTa-Large \small{(435M)} &  30.1 & 95.5 & 24.7 & -1.0 \\
% UnifiedQA & 60.8 & 91.4 & 17.9 & -0.3 \\
\midrule
  \multicolumn{5}{c}{\textit{GPT-3 Few-Shot Prompting}} \\
  \midrule
  0-shot & 60.5 & 43.2 & 3.7 & 4.4 \\
  % RACE Prompt& 95.4 & 51.7 & 7.5 & 2.7 \\
  BBQ Balanced & 96.8 & 76.0 & 2.4 & 1.5 \\
  % BBQ DisAmbig & 24.4 & 97.9 & 22.0 & 1.9 \\
  BBQ Ambig-Neutral & 99.9 & 13.2 & 0.0 & -3.5 \\
  BBQ Ambig-Pro-Bias & 2.6 & 97.3 & 24.7 & 3.2 \\
  BBQ Ambig-Anti-Bias & 2.5 & 97.0 & 23.6 & 3.1 \\
 \bottomrule
\end{tabular}}
 \caption{Results on the BBQ dataset. 
 % all results are the macro-average across all different bias categories.
 For GPT-3 prompting, apart from the zero-shot result, others use 8-shots. 
 For accuracy (Acc), higher value is better; for bias score, lower magnitude is better. The balanced prompt best
 trades-off accuracy and bias for GPT-3.  
 } 
 \label{tab:bbq}
\end{center}
% \vspace{-1mm}
\end{table}

\begin{table}[t]
\small
\begin{center}
\setlength{\tabcolsep}{2mm}{
\begin{tabular}{ l  c c c  c }
\toprule
  Prompt & Ambig Acc$_{\uparrow}$ & DisAmbig Acc$_{\uparrow}$ & Ambig Bias Score$_{\downarrow}$ & DisAmbig Bias Score$_{\downarrow}$ \\
  \midrule
 Before Intervention & 2.6 & 97.3 & 24.7 & 3.2 \\
 After Intervention & 96.6 & 51.5 & 1.9 & 3.8 \\
 \bottomrule
\end{tabular}}
 \caption{The impact of adding natural language intervention to a biased prompt on GPT-3. Adding an instruction leads the model to make more neutral predictions and reduce bias scores.} 
 \label{tab:bbq_intervention}
 % \vspace{-1mm}
\end{center}
\end{table}

%\vspace{-2mm}
\subsection{Broader Social Dimensions: BBQ}
%\vspace{-2mm}

\paragraph{Dataset} We now explore additional social dimensions using BBQ~\citep{BBQ}, which tests
social biases against people from nine protected classes (age, disability status, gender identity, nationality, physical appearance, race, religion, socio-economic status, sexual orientation). 
BBQ examples are in sets of four multiple-choice questions. 
Two questions are ambiguous---the context lacks evidence to point to an answer.  
Two other questions in each set have a context that points to an unambiguous answer: the model should choose the correct answer rather than abstaining.
Each question has three options: a pro-bias answer that supports the stereotype, an anti-bias answer that counters the stereotype, and a neutral answer (\textit{e.g.}, ``\textit{Cannot be determined.}''). 
When reporting results, we report:
1) accuracy on ambiguous and unambiguous questions (higher is better); 
2) bias scores on ambiguous and disambiguated questions (smaller scale is better). 
Intuitively, the bias score measures the frequency of the model predicting a pro-bias answer when it makes a non-unknown prediction, where 0\% means no bias and 100\% means always following the bias.  
% \jbgcomment{pro-bias is not clear for BBQ dataset.  Perhaps add something like: "For ambiguous questions, there is a pro-bias answer that\dots"}

% \paragraph{Baselines} We take RoBERTa, DeBERTa, and UnifiedQA-11B results from \cite{BBQ} as baseline references. RoBERTa and DeBERTa are finetuned on RACE~\citep{Lai2017RACELR}, another multiple-choice QA datasets sourced from middle and high school reading comprehension exams. For UnifiedQA, it is trained on eight QA datasets and evaluated in the multiple-choice format similar to RACE. 

% \jbgcomment{The different prompts are not clear, and don't just say "different" in the paragraph.  What's the takeaway from this experiment?}
\paragraph{Answer Distribution Affects Biases} 
We sample the same number of demo examples with different answer distributions (Table~\ref{tab:bbq}).
With zero-shot inference, GPT-3's low bias is balanced by low
accuracy.
A balanced prompt---where we sample both ambiguous and
disambiguated questions with pro-bias and anti-bias answers---increases accuracy while
maintaining low bias scores.
Next, when we only present ambiguous questions with neutral answers in
the prompt, GPT-3 can easily pick up this pattern to always predict
the neutral answer, resulting in extremely low bias scores and high
accuracy on ambiguous questions, but sacrificing accuracy on the
disambiguated questions.
Lastly, we experiment with using biased prompts where we
only sample ambiguous questions in the prompt, and we set their
answers to either all bias-supporting or bias-countering. As shown in
the last two rows in Table~\ref{tab:bbq}, both cases lead to a biased
model that gets high bias scores on ambiguous questions.  This
shows that when the demo examples are biased towards certain
demographic groups, GPT-3 will pick up such biases, and thus one
should prefer a balanced prompt in downstream applications.

% \paragraph{Category Breakdown}
% We also break down the accuracy and bias scores of using different prompts in Table~\ref{tab:bbq_breakdown} by the different bias categories. We observe that there can be large differences across different categories. Moreover, we underlined the categories from which the demo examples come, and we observe that having same-category demos in the prompt does not correlate with the bias scores. For instance, we have bias-supporting examples from the Nationality category in the Ambig-Pro case but the bias score remains low, while the bias score for the Physical Appearance and Disability categories becomes much higher even when the biased examples are not from these categories. 

\paragraph{Natural Language Intervention Mitigates Biases}
Lastly, we explore the possibility of using natural language instructions to guide the predictions of GPT-3 towards better fairness. We use the Ambig-Pro-Bias prompt and prepend the additional instruction: ``\textit{We should treat people from different socioeconomic statuses, sexual orientations, religions, races, physical appearances, nationalities, gender identities, disabilities, and ages equally. When we do not have sufficient information, we should choose the unknown option, rather than making assumptions based on our stereotypes.}'' to every test question. In Table~\ref{tab:bbq_intervention}, after adding this intervention, GPT-3 prefers the neutral answer much more often, leading to a much higher accuracy on the ambiguous questions, and at the same time significantly reducing the bias scores. This shows that GPT-3 is sensitive to such natural language intervention. This is in  contrast with smaller language models such as RoBERTa~\citep{Zhao2021EthicalAdviceTD}, which is more rigid. This finding offers a new way for effectively reducing social biases. 

\paragraph{Takeaway} ($i$) Demographic distribution of answers has huge impact on models' biases, sampling balanced prompt best reduces biases. ($ii$) Randomly shuffling the demos leads to smaller biases than putting all pro-bias or anti-bias examples in the end. ($iii$) Specifying intended model behaviors such as being fair via instructions in the prompt can effectively guide model predictions.

\begin{wraptable}{R}{6.5cm}
\vspace{-3mm}
\setlength{\tabcolsep}{1mm}{
\begin{tabular}{ l c  c  c }
\toprule
& Acc$_{\uparrow}$ & ECE$_{\downarrow}$ & Brier$_{\downarrow}$ \\
   \midrule
  \multicolumn{4}{c}{NQ} \\
  \midrule
DPR-BERT \small{(110M)} &  36.1 & 29.4 & 33.5 \\
GPT-3 LM Prob & 40.5 & 18.9 & 23.3 \\
GPT-3 Self-Con & 40.2 & 14.3 & 20.1 \\
  \midrule
  \multicolumn{4}{c}{TriviaQA (TQA)} \\
  \midrule
GPT-3 LM Prob & 73.8 & 3.8 & 15.9 \\
GPT-3 Self-Con & 73.2 & 11.9 & 16.5 \\
   \midrule
  \multicolumn{4}{c}{HotpotQA (HQA)} \\
  \midrule
GPT-3 LM Prob & 29.8 & 25.0 & 23.5 \\
GPT-3 Self-Con & 28.5 & 20.7 & 19.9 \\
   \midrule
  \multicolumn{4}{c}{Different Prompts on NQ w/ LM-Prob} \\
  \midrule
GPT-3 2-shot & 37.0 & 11.7 & 20.8 \\
GPT-3 4-shot & 38.3 & 13.4 & 21.0 \\
GPT-3 8-shot & 38.8 & 24.4 & 25.5 \\
GPT-3 16-shot & 40.5 & 18.9 & 23.3 \\
GPT-3 64-shot & 42.8 & 13.4 & 22.1 \\
% \midrule
%  \multicolumn{4}{c}{Different Models on TQA w/ LM-Prob} \\
% \midrule
% Code-Davinci-002 & 73.8 & 3.8 & 15.9 \\
% Text-Davinci-001 &  50.5 & 28.9 & 29.9 \\
% Text-Curie-001 &  23.6 & 40.3 & 32.7 \\
\midrule
  \multicolumn{4}{c}{OOD Prompts w/ LM-Prob} \\
\midrule
TQA i.i.d. Prompt & 73.8 & 3.8 & 15.9 \\
NQ Prompt on TQA & 73.0 & 1.6 & 15.2 \\
DPR-BERT NQ $\rightarrow$ TQA & 33.1 & 33.1 & 35.2 \\
\midrule
HQA i.i.d. Prompt & 29.8 & 25.0 & 23.5 \\
NQ Prompt on HQA & 27.7 & 24.1 & 25.2 \\
DPR-BERT NQ $\rightarrow$ HQA & 23.6 & 45.7 & 42.4 \\
 \bottomrule
\end{tabular}
\caption{Accuracy, ECE, and Brier scores of GPT-3 and the DPR-BERT baseline. GPT-3 is better calibrated than supervised DPR-BERT on
  both in-domain and OOD settings.}
% \vspace{-1mm}
\label{tab:calibration_ece}
}
\end{wraptable} 

% \zhe{some tables may hard to read. can we indicate explicitly in the metrics of tables like table 5 and 6 etc, higher better or lower better, w/o digging into the captions? for example, by adding upper and down arrows.}

% \jbgcomment{\huge Don't use vspace hacks!  This can get your paper desk rejected.}

%\vspace{-2mm}
\section{Facet 3: Uncertainty Calibration}
%\vspace{-2mm}

No language model can ever be perfect, and to safely use these imperfect models, users must decide when to \textbf{trust} model predictions to avoid mistrusting wrong predictions, especially in high-stake settings. 
This requires another facet of reliability - uncertainty calibration: providing confidence scores for each model prediction that accurately reflects the likelihood of the predicted answer being correct. 

%\vspace{-2mm}
\subsection{Evaluation Setup}
%\vspace{-2mm}
% While a series of prior works focused on calibrating GPT-3 style autoregressive language models on multiple-choice questions, 
\paragraph{Experiment Setup}
We study the setting of free-form answer generation:
given a test question, we prompt the model to generate an answer string and obtain its confidence score (more below), and we evaluate the correctness of the generated answer based on exact match with the gold answer. 
We experiment with three QA datasets: NQ, TriviaQA, and HotpotQA. In all cases, we adopt the closed-book setting (i.e., no additional evidence passages). 
We focus on intrinsic calibration results: using raw confidence scores rather than post-hoc calibration, which requires an additional dev set for parameter-tuning. 
We report the standard calibration metric expected calibration error (ECE), the reliability diagram,\footnote{In Appendix Figure~\ref{fig:calibration_plots}.} and selective prediction results where we rank all predictions by their confidence and see if the accuracy of the most confident predictions is significantly higher than the average accuracy. Because of ECE's known flaws due to its bucketing mechanism~\citep{Si2022RevisitingCF}, so we also report the Brier score~\citep{Brier1950VERIFICATIONOF}. 
Our baseline is a supervised QA
model---DPR-BERT~\citep{Si2022RevisitingCF}---with a dense passage
retriever~\citep[DPR;][]{Karpukhin2020DensePR} to feed the top
passages into a BERT reader model for answer extraction. We follow
their joint calibration setup for scoring predictions of DPR-BERT.

%\vspace{-2mm}
% \subsection{Confidence Scoring}
%\vspace{-2mm}

\paragraph{Confidence Scoring}
We compare two ways of estimating confidence for GPT-3 predictions. \textbf{LM-Prob}: the (normalized) language model probability, also equivalent to the reciprocal of perplexity, is $Conf \equiv P(w_1 w_2 \dots w_n)^{\frac{1}{N}}$ where $w_1 w_2 \dots w_n$ are the generated tokens in the answer. \textbf{Self-Con}: We also explore using self-consistency~\citep{Wang2022SelfConsistencyIC} to obtain confidence measures. Following~\cite{Wang2022SelfConsistencyIC}, during decoding we set a high temperature value (0.7) and sample 10 times for a set of different predictions. Among all the generated answers, we take the most frequent answer as the final prediction and its frequency as the confidence score.  

\begin{table}[t]
\small
\begin{center}
\setlength{\tabcolsep}{2mm}{
\begin{tabular}{ l c c c c c }
\toprule
& DPR-BERT NQ & LM-Prob NQ & Self-Con NQ & LM-Prob TriviaQA & LM-Prob HotpotQA \\
\midrule
100\% & 36.1 & 40.5 & 40.2 & 73.8 & 29.8 \\
% 90\% & 38.0 & 43.7 & 44.3 & 78.3 & 32.7 \\
% 80\% & 39.5 & 46.8 & 48.7 & 81.7 & 36.0 \\
% 70\% & 40.6 & 50.2 & 53.1 & 84.1 & 39.7 \\
% 60\% & 41.2 & 53.7 & 57.8 & 86.5 & 43.5 \\
50\% & 41.9 & 58.8 & 62.0 & 88.5 & 47.6 \\
% 40\% & 43.3 & 63.3 & 66.0 & 90.5 & 52.1 \\
% 30\% & 46.1 & 70.2 & 71.2 & 92.5 & 56.5 \\
% 20\% & 49.2  & 77.4 & 74.7 & 93.7 & 61.6 \\
10\% & 60.1 & 83.1 & 77.0 & 95.4 & 68.1 \\
 \bottomrule
\end{tabular}}
 \caption{Selective prediction results. All numbers represent the accuracy (EM) at the corresponding coverage thresholds. For example, 100\% means performance on the entire test set while 10\% means the performance on the most confident 10\% predictions.
 Both LM-Prob and Self-Con allow effective selective prediction with high accuracy on the most confident subsets. More results in Appendix~\ref{sec:more_calibration}. 
 } 
 % \jbgcomment{Make sure that every caption has a takeaway}
 \label{tab:selective_prediction}
 % \vspace{-3mm}
\end{center}
\end{table}

% \jbgcomment{You don't formally define what ``selective prediction is''}

%\vspace{-2mm}
\subsection{Results}
%\vspace{-2mm}
While still imperfect, GPT-3 (with either LM-Prob or Self-Con) is better calibrated than supervised DPR-BERT (Table~\ref{tab:calibration_ece}). Most calibration errors come from overconfidence where the predictions' confidence is higher than expected accuracy. Interestingly, while increasing the number of examples in the prompt improves accuracy, the calibration does not improve. For example, the 2-shot accuracy is 5.8 points worse than 64-shot but better calibrated. 
Moreover, while OOD transfer is a challenge for supervised models' calibration (tends to be overconfident on OOD test sets), GPT-3 has similar calibration regardless of the source of examples.
% (bottom-left plot in Figure~\ref{fig:calibration_plots}). 

The selective prediction results show confidence scores can rank model predictions (Table~\ref{tab:selective_prediction}): the most confident predictions have much higher accuracy. 
Moreover, GPT-3's confidence scores are more discriminative.
For example, while the average accuracy 
% \jbgcomment{this is a case where you simply cannot use "performance": is this calibration or accuracy} 
on NQ is similar between GPT-3 and DPR-BERT, the top 10\% predictions get an accuracy of 83.1\% while for DPR-BERT it is only 60.1\%.  
Such selective prediction can be very useful in practical settings, for example, we only trust the most confident predictions from the model and ask humans to verify the rest, making the use of GPT-3 more reliable.  

\textbf{Takeaway} ($i$) Language model probability and self-consistency frequency can produce better calibration on GPT-3 than a supervised DPR-BERT model, especially on OOD test sets. ($ii$) Increasing the number of demos in the prompt improves accuracy but not necessarily calibration. ($iii$) We can perform effective selective prediction based on GPT-3 confidence scores. 

% The reliability diagrams in Figure~\ref{fig:calibration_plots} show that in most cases the calibration errors come from overconfidence where the predictions' confidence is higher than the  expected accuracy. 

% \zhe{how do all these results relate to Prompting GPT-3 to be More Reliable? It seems that it only shows the behaviors of GPT-3?}

%\vspace{-2mm}
\section{Facet 4: Factuality Via Knowledge Updating} \label{sec:knowledge_updating}
%\vspace{-2mm}

Although large language models store vast knowledge in their parameters~\citep{Petroni2019LanguageMA}, the model is sometimes wrong or out of date, rendering them unreliable for knowledge-intensive tasks. In this section, we improve this factuality aspect of reliability by improving the prompting methods.

% \vspace{-1mm}
\subsection{Memorization vs Updating}
% \vspace{-1mm}

The larger a model, the more it can memorize~\citep{Carlini2022QuantifyingMA}, this  raises the concern of whether large models like GPT-3 can \textbf{forget} memorized knowledge when needed and \textbf{update} its knowledge. 
% Toward this goal, we evaluate the effectiveness of \textbf{in-context knowledge updating}. 

\begin{wraptable}{R}{7.9cm}
\vspace{-2mm}
\setlength{\tabcolsep}{1.3mm}{
\begin{tabular}{ l c c c }
\toprule
& Retain$_{\downarrow}$ & Update$_{\uparrow}$ & Other$_{\downarrow}$  \\
\midrule
  \multicolumn{4}{c}{\textit{NQ} with \textit{Code-Davinci-002}}\\
\midrule
T5 \small{(770M, supervised)} & 20\% & 33\% & 47\%  \\
GPT-3 & 4.5\% & 85.4\% & 10.2\%  \\
\midrule
\multicolumn{4}{c}{\textit{SQuAD} with \textit{Code-Davinci-002}} \\
\midrule
GPT-3  & 7.1\% & 84.8\% & 8.1\%  \\
\midrule
\multicolumn{3}{c}{\textit{NQ} with different GPT-3 models} \\
\midrule
\textit{Text-Davinci-001} (175B) & 7.2\% & 57.9\% & 34.9\% \\
\textit{Text-Curie-001} (6.7B) & 14.8\% & 40.0\% & 45.2\% \\
 \bottomrule
\end{tabular}}
 \caption{In-context knowledge updating results for memorized answers
   in NQ and SQuAD. When giving counterfactual examples in the prompt,
   GPT-3 updates its answers around 85\% of the time, much higher
   compared to the supervised model. Moreover, larger models are
   better at in-context knowledge updating.}   
 \label{tab:memorization}
 % \vspace{-1mm}
\end{wraptable}

\paragraph{Experiment Setup} Our evaluation setup is inspired by~\cite{Longpre2021EntityBasedKC}, who reason about counterfactual scenarios.  Specifically, we sample 36K and 18K questions from NQ and SQuAD's training splits (respectively, using the splits provided by MRQA).
We use 16 demo examples from each dataset as the prompt for closed-book QA first. 
We assume that if GPT-3 gets the answer to the question right in the closed-book setting, then it has already memorized that piece of knowledge.
We keep the set of questions where GPT-3 got right in the closed-book setting (for NQ, 21188 questions; for SQuAD, 7035 questions), and for these questions, we append a counterfactual passage supporting an alternative answer.
We construct these counterfactual using the entity-swap from \cite{Longpre2021EntityBasedKC}: for each question, take its gold passage and replace the gold answer entity with another entity with the same type sampled from the same QA corpus. 
% \jbgcomment{This would be stronger with things that can change, e.g., use the classifier to predict timely questions and focus on those.}
%
After such entity substitution, the counterfactual passages support the substituted answer instead of the original answer. Our expectation is that the model should generate this updated answer given this counterfactual passage, instead of its original memorized answer. 
%
% \paragraph{Prompt Design}
% \jbgcomment{I'm not sure this detail needs to be in the main paper}
% We compare several different prompt designs as detailed below, for all cases, we randomly sample 16 demo examples as the prompt:
We randomly sample 16 demo examples as the prompt and 
% 1) $\langle$\textit{Q, A}$\rangle$ : We use the original question-answer pairs in the prompt; 
% 2) $\langle$\textit{P, Q, A}$\rangle$ : We use the original passage-question-answer triples in the prompt (i.e., the answer in the passage remains the original gold answer);
% 3) $\langle$\textit{Q, A'}$\rangle$ :  We use the question-answer pairs, but with the substitution entities as gold answers in the prompt;
 % $\langle$\textit{P', Q, A'}$\rangle$ : 
 we use triples of the answer-substituted passage, the question, and the substitution answers ($\langle$\textit{P', Q, A'}$\rangle$) in the prompt to specify the task of performing reading comprehension based on the passage. 

\paragraph{Measuring How Well can GPT-3 Update its Knowledge} There are three possible outcomes: 1) the model retains the memorized answer; 2) the model predicts the updated answer (i.e., the substitution entity in the counterfactual passage); 3) the model predicts some other answer.
We measure the proportion of those outcomes and hope models to update answers more often. 
% as well as the memorization ratio~\citep[MemRatio]{Longpre2021EntityBasedKC}: $\frac{\textrm{Retain}}{\textrm{Retain} + \textrm{Update}}$, how often the model retains the memorized answer as opposed to updating. 
For a baseline, we include results from \cite{Longpre2021EntityBasedKC}: a fine-tuned T5 reader---trained on NQ and NewsQA---model with a DPR retriever. 

\paragraph{Results}
As shown in Table~\ref{tab:memorization}, 
we find that 
% the prompt design has a big impact on the knowledge updating behavior. In particular, showing only the original passage-question-answer triples ($\langle$\textit{P, Q, A}$\rangle$) still causes high memorization ratios, however, 
when prompting with counterfactual triples ($\langle$\textit{P', Q, A'}$\rangle$), GPT-3 can update about 85\% of the time, much higher than the supervised baseline  (Table~\ref{tab:memorization}). 
% with much lower memorization ratios than a supervised model. 
Comparing \textit{Text-Davinci-001} and \textit{Text-Curie-001},
the larger model also updates better to new answers in counterfactual passages.

%\vspace{-2mm}
\subsection{Retrieval-Augmented Open-Domain QA}
%\vspace{-2mm}

Large language models can answer closed-book QA from  the model's stored knowledge~\citep{Roberts2020HowMK}. 
However, a prompt can judiciously add more relevant information especially given our findings from the previous section that GPT-3 can update its knowledge with information in the prompt. We thus explore improving factual QA via retrieval-augmented prompts. 
%
% We challenge the underlying assumption that the factual knowledge from the pretraining corpora is all captured in the parametric memory, and study where GPT-3 can benefit from retrieved knowledge added to the prompt. 
% We experiment with adding retrieved passages from a retriever to GPT-3 and see whether it still brings additional gains compared to the closed-book baseline. 

% Traditional people tackle the problem of open-domain question answering with a retriever-reader pipeline where a retriever (either sparse or dense) is used to retrieve the most relevant passages from knowledge bases like Wikipedia, and then a reader model is used to extract answers from these passages~\citep{Chen2017ReadingWT,Karpukhin2020DensePR}. However, with large language models, it is now also possible to perform closed-book QA on these tasks without using a retriever, relying solely on the model's stored parametric knowledge~\citep{Roberts2020HowMK}. 

% We challenge the underlying assumption that the factual knowledge from the pretraining corpora is all captured in the parametric memory. In order to do so, we experiment with adding retrieved passages from a retriever to GPT-3 and see whether it still brings additional gains compared to the closed-book baseline. 

\paragraph{Approach} We use the unsupervised Contriever model~\citep{Izacard2021UnsupervisedDI}: for a test question,  retrieve the top passages from the  
% Dec. 20, 2018 \jbgcomment{why this date?  It doesn't sound very up-to-date} 
Wikipedia dump, concatenate them, and prepend them to the test question. 
Since the context is length-limited, we only prepend retrieved passages to the test question, not the demo examples, so the demo examples are only in the form of question-answer pairs. We compare this retriever-augmented approach with a closed-book baseline where we do not add the retrieved passages in the prompt. The demo examples used for both the retrieval-augmented prompting and closed-book prompting are exactly the same.

\begin{table}
\small
\begin{center}
\setlength{\tabcolsep}{3.5mm}{
\begin{tabular}{ l  l l l }
\toprule
 & NQ & TriviaQA & SQuAD \\
  \midrule
  DPR-BERT (supervised) & 41.5 & 56.8 & 24.1 \\
  Atlas-11B (64-shot) & 42.4 & 74.5 & -- \\
  \midrule
GPT-3 Closed-Book & 40.6 & 73.6 & 20.2 \\
+ Contriever top-5 & 43.3 (61.8\%) & 75.6 (69.6\%) & 31.7 (48.8\%) \\
+ Contriever top-10 & 44.2 (70.5\%) & 76.0 (75.1\%) & 34.0 (57.7\%) \\
% \midrule
% Closed-Book $\cup$ Contriever top-5 & 49.4 & 79.3 & 35.8 \\
% Closed-Book $\cup$ Contriever top-10 & 50.2 & 79.6 & 37.6 \\
 \bottomrule
\end{tabular}}
 \caption{GPT-3 16-shot prompting results on open-domain QA datasets. We use Exact Match as the metric. For Contriever results, we additionally show the retriever's recall in brackets.  
 % The last two rows show the oracle ensemble of the closed-book model and the retrieval-augmented model.
 Adding retrieval to GPT-3 consistently improves QA accuracy. 
 }  
 \label{tab:retrieve_odqa}
 \vspace{-2mm}
\end{center}
\end{table}

\paragraph{Results}
Adding retrieved passages into the prompt consistently boosts GPT-3 performance on all three open-domain QA datasets (Table~\ref{tab:retrieve_odqa}), with particularly large gains on SQuAD (possibly because answers in SQuAD are spans from Wikipedia passages rather than free-form answers).  Moreover, having better recall for retrieval gives better performance.

\vspace{-1mm}
\subsection{Reasoning-Augmented Multi-Hop QA}
%\vspace{-2mm}

% \jbgcomment{This seems like the weakest section.  What is new here that hasn't been seen in previous papers?  Why does it belong here?  What is the contribution here?}

% While the previous subsections showed that GPT-3 is able to adapt to new information in the prompt, these experiments are focused purely on factual knowledge. In this subsection, we explore \jbgcomment{explore is a weak word, be more specific} another form of knowledge: reasoning chains.

The above experiments demonstrate the effectiveness of ensuring GPT-3's factuality via in-context knowledge updating; however, it is mostly constrained on simple single-hop factual questions. In real-world applications, many user queries are multi-hop - they require multiple steps of reasoning over factual knowledge. Ensuring factuality in multi-hop questions involves additional challenges: models may fail because they derive the reasoning steps wrongly. To tackle this more challenging multi-hop setting, we study whether it is possible to improve GPT-3's multi-hop reasoning by incorporating human-written question decomposition in the prompt. 

\textbf{HotpotQA and Decomposed Sub-Questions}
We use the HotpotQA dataset~\citep{Yang2018HotpotQAAD} for our experiments, which consists of multi-hop questions that require at least two steps of reasoning. 
% The focus of our experiment is whether incorporating an explicit form of question decomposition (written by humans) can help GPT-3 answer multi-hop questions. 
We use the question decomposition from \cite{Tang2021DoMQ}, where HotpotQA questions are annotated as decomposed (single-hop) sub-questions with corresponding intermediate answers. 
% We evaluate several prompting methods both on the overall answer accuracy as well as the accuracy on the sub-questions.

\begin{table}
\small
\begin{center}
\setlength{\tabcolsep}{3mm}{
\begin{tabular}{ l c c c }
\toprule
& Overall & Sub-Q1 & Sub-Q2 \\
\midrule
Standard Prompting & 18.0 / 28.1 & 40.1 / 49.6 & 43.3 / 58.4 \\
CoT & 25.2 / 35.2 & 30.3 / 37.4 & -- \\
CoT + Human Sub-Q1 & 30.0 / 42.3 & 44.2 / 54.1 & -- \\
CoT + Human Sub-Q1 + Gold Sub-A1 & 44.3 / 59.0 & -- & -- \\
 \bottomrule
\end{tabular}}
 \caption{Results on HotpotQA as well as the decomposed sub-questions (we report EM / F1). Incorporating decomposed sub-questions in the prompt makes the model adjust its follow-up step predictions and significantly improves the overall answer accuracy.}   
 \label{tab:hotpot_cot}
 \vspace{-3mm}
\end{center}
\end{table}

\textbf{Baseline: Chain-of-Thought Prompting}
Chain-of-Thought (CoT) prompting~\citep{Wei2022ChainOT} is a new prompting method tailored to multi-step questions, which we adopt in our experiments as a baseline, where we provide human-written reasoning steps for all demo examples to induce similar reasoning on test examples. 
We measure accuracy of GPT-3's final answer predictions on HotpotQA (Overall) as well as on the decomposed single-hop sub-questions. 
From the first row of Table~\ref{tab:hotpot_cot}, we see that standard prompting achieves higher accuracy on the single-hop sub-questions than the entire multi-hop questions as expected. 
%
% much higher performance when prompted to solve the single-hop sub-questions as compared to directly generating the final answer to the overall multi-hop question. When performing Chain-of-Thought prompting, we do see a non-trivial improvement on the overall QA performance (EM 18.0 to 25.2). 
CoT generates the entire reasoning chain along with its decomposed sub-questions and the intermediate answers to sub-questions, 
where the accuracy on the multi-hop questions is higher than standard prompting (second row of Table~\ref{tab:hotpot_cot}). 
% This suggests that a potential bottleneck when solving multi-hop questions is how to get intermediate steps and answers right in order to avoid cascading errors. 

\textbf{Incorporating Human Decomposition}
% To improve the intermediate step accuracy, 
Instead of relying on GPT-3 itself to generate reasoning chains,
we add the human-written question decomposition into the prompt. When adding the human-written  sub-questions for the first step of reasoning (second last row of Table~\ref{tab:hotpot_cot}), we see a clear improvement in both the overall multi-hop QA accuracy as well as the sub-question accuracy. Moreover, when we further add the human-written QA pair of the first decomposed question in the reasoning chain (last row of Table~\ref{tab:hotpot_cot}), there is an even larger performance gain on the multi-hop QA performance. This shows that GPT-3 is able to adapt to the question decomposition information from humans and deduce the subsequent reasoning steps to eventually obtain the correct answers, offering better control and reliability. 

\textbf{Takeaway} ($i$) Adding retrieved evidence passages can improve GPT-3 performance on factual QA. ($ii$) GPT-3 can update its knowledge when provided passages conflicting with its memorized knowledge. ($iii$) Incorporating human-written question decomposition corrects the reasoning chains of GPT-3 and improves performance on multi-hop QA.

\vspace{-1mm}
\section{Conclusion} \label{sec:conclusion}
\vspace{-1mm}

Our work systematically studies the reliability of GPT-3 from four key facets: generalizability, fairness, calibration, and factuality. We develop effective prompting strategies to make GPT-3 outperform supervised models by large margins on these facets. 
Our work reveals new insights of LLMs and provides practical recommendations for users of GPT-3. 
We hope our work can inspire more future work to: (1) examine more facets of reliability, such as avoiding harmful generations; (2) apply the prompting methods in this paper to more real-world applications, such as incorporating human feedback for collaborative multi-step planning; (3) further explore more effective prompting strategies to improve reliability, such as post-hoc calibration on language model probabilities. 
% Moreover, with more carefully designed prompts, we show GPT-3's capability to adapt to new information in the prompt, such as natural language interventions, knowledge updates, and reasoning steps. We believe that our findings can serve as useful guidelines for practitioners using LLMs like GPT-3, and inspire more work to improve different aspects of reliability. 

\clearpage
\section*{Ethical Statement}

\paragraph{Ethical Use of GPT-3}
The goal of this project is to avoid the potential harm of GPT-3 and all of our GPT-3 experiments are motivated to better study and improve reliability. We believe our experiments and findings can improve the reliability and allow safer use of the model. In particular, our section on social biases and fairness is a key aspect of the ethical use of GPT-3. We presented evidence that the model exhibits biased predictions, especially when the demo examples in the prompt have a skewed demographic distribution. Although we explored ways of mitigating these biases, the model is still far from perfect, and there is much more work needed to further improve its fairness. We take our work as an initial step towards more ethical use of GPT-3. 

\paragraph{Limitations of This Work}
We note several limitations of this work and suggest a list of open questions for future work. 

\begin{itemize}
    \item \textbf{Other reliability facets}: In this work, we covered four key facets of reliability, but there are surely other facets that we may have missed. For example, combatting adversarial examples identified via human or AI red-teaming~\citep{Ganguli2022RedTL,Branch2022EvaluatingTS,Perez2022RedTL}, detecting and handling malicious prompts such as prompt injection~\footnote{\url{https://simonwillison.net/2022/Sep/12/prompt-injection/}}, and avoiding toxic and hallucinated generations~\citep{Gehman2020RealToxicityPromptsEN,Gao2022AttributedTG}.

    \item \textbf{Methods for improving reliability}: Although we have taken initial steps and discovered some effective prompting strategies for these reliability facets, readers should not take this work as evidence that GPT-3 is already reliable and ready for deployment. In fact, our experiments indicate ample room for further improvement, for example in reducing social biases and improving calibration. We hope this work inspires more future work that develops more effective strategies to make LLMs reliable. 

    \item \textbf{Analysis to understand model behaviors}: While we have found interesting properties of GPT-3, it remains unclear what exactly caused these behaviors. For example, if the small generalization gap due to the use of prompting, or the training data, or the training objectives or model architecture? When GPT-3 is sensitive to the prompt in debiasing, is it triggered by certain keywords or phrases? Why ordering anti-bias examples at the end of the prompt does not lead to the recency bias~\citep{Zhao2021CalibrateBU} but rather still incurs strong biases against minority groups? Can we attribute model behaviors to the pretraining data or interpret model attention patterns? These analysis can potentially help us better understand how and why prompting works and therefore allow us to better leverage LLMs.
\end{itemize}

% \subsection*{Broader Impact}

% \section*{Ethical Statement}

\section*{Acknowledgment}
We thank Jason Phang, Ziyi Yang, Dan Friedman, Sewon Min, Jieyu Zhao, He He, Alicia Parrish, Chen Zhao, Shi Feng, Han Guo, Weijia Shi, Jungo Kasai, Xi Ye, Su Lin Blodgett, Trista Cao, Ekin Akyürek, Leo Boytsov, Aishwarya Kamath, Weijia Xu, Yankai Lin, Xiaozhi Wang, Zhengyan Zhang, and many other friends from UMD CLIP and the Azure AI team at Microsoft for their helpful discussion and feedback. 

\clearpage

\bibliography{iclr2023_conference}
\bibliographystyle{iclr2023_conference}

\clearpage
\appendix
\section*{Appendix}

\section{More Related Work}
\label{sec:related_work}

% Introduce and Motivate each aspect; give a short summary of all major evaluation setups and findings. 

% In this section, we outline our proposed reliability testing framework designed for LLMs. The framework  consists of four core factors, and we provide motivation and an introduction for each of them. 

\paragraph{Robustness to Distribution Shifts.} Machine learning models are known to overfit their training distribution and often suffer performance degradation when the test distribution differs from the training distribution. In the case of language models, various forms of distribution shifts have been studied. For example, domain shifts pose great challenges for LLMs on question answering~\citep{Talmor2019MultiQAAE,Fisch2019MRQA2S} and text classification~\citep{Hendrycks2020PretrainedTI,Arora2021TypesOO}; various forms of adversarial attacks can break LLMs even by just strategic synonym substitution, paraphrase, or distractor insertion~\citep{Jin2020IsBR,Li2020BERTATTACKAA,Ribeiro2018SemanticallyEA,Si2019WhatDB,Si2021BenchmarkingRO,Jia2017AdversarialEF,Si2020BetterRB}; LLMs have been shown to exploit shortcuts or spurious correlations in the training data which fail on counter test examples~\citep{HANS,PAWS,Poliak2018HypothesisOB,Gururangan2018AnnotationAI}; LLMs also fail on new compositional structures that are not observed during traing~\citep{Kim2020COGSAC,Keysers2020MeasuringCG,Bogin2022UnobservedLS}. In real-world settings, various forms of distribution shifts can happen and reliable models should perform well even when encountering such out-of-distribution (OOD) examples. Intuitively, in-context few-shot prompting should suffer less OOD degradation since the pretrained parameters are preserved, unlike the case of supervised finetuning. We perform a series of empirical evaluations on domain shift, curated challenge sets, and spurious correlation to validate this hypothesis.

\paragraph{Bias and Fairness.}
Language models producing toxic or biased content can cause severe harm especially to the groups being biased against~\citep{Bender2021OnTD}. A series of benchmarks have been developed to show that LLMs can generate toxic outputs~\citep{Gehman2020RealToxicityPromptsEN}, contain gender biases~\citep{Rudinger2018GenderBI,Zhao2018GenderBI} and other categories of social biases~\citep{Nangia2020CrowSPairsAC,Nadeem2021StereoSetMS,BBQ}, perform poorly against minority demographic groups~\citep{Koh2021WILDSAB,Harris2022ExploringTR} or dialectical variations~\citep{Ziems2022VALUEUD,Tan2020ItsMT}. Ideally, LLMs should not exhibit biased behaviors and not discriminate against any group. While many of these evaluations focus on evaluating the internal representation of LLMs in a zero-shot setting or evaluating the biases on specific downstream applications in a supervised setting, it remains unclear how these biases change under different prompting schemes in the few-shot setting, which will be the focus of our analysis. A closely related work is \cite{Lucy2021GenderAR} which study representation biases in GPT-3 generated stories. We instead evaluate on the downstream tasks of coreferece resolution and question answering. Apart from few-shot prompting, \cite{Solaiman2021ProcessFA} proposed a general method to align language models with human values, but it involves expensive iterative training.

\paragraph{Uncertainty Calibration.}
No model can ever be perfect, and so it is crucial for users to be able to identify model mistakes, especially in high-stage settings where trusting wrong model predictions can cause severe harm. One important way to help identify wrong model predictions is by obtaining well-calibrated confidence scores for model predictions. By definition, a calibrated confidence (probability) score should match the expected accuracy of the prediction~\citep{Platt1999ProbabilisticOF,Naeini2015ObtainingWC,Guo2017OnCO}. In this way, users can put more trust in highly-confidence predictions and discard low-confidence predictions. While various methods have been proposed to obtain confidence scores and perform post-hoc calibration for language models~\citep{Jiang2021HowCW,Desai2020CalibrationOP,Ye2022CanEB}, they are mostly focused on classification settings rather than free-form generation, which is more common for the use of GPT-3. In this work, we explore two simple (but surprisingly effective) ways of obtaining confidence scores for GPT-3's generated answers and we analyse the impact of scaling as well as prompt design. 
For studying calibration of GPT-3 style LLMs, \cite{Lin2022TeachingMT} explore the idea of expressing uncertainty in verbal words but is restricted to math questions. \cite{Mielke2022ReducingCA} study linguistic calibration on conversational models. 
\cite{Kadavath2022LanguageM} study adopting a multiple-choice setting in which case obtaining a confidence score is much easier (since the model only needs to predict one token to indicate which option to choose rather than generating the entire answer string). We differ from them in: 1) we focus on obtaining probabilistic confidence scores rather than verbal uncertainty expressions; 2) we study the more general and realistic free-form answer generation setting; and 3) we do not involve finetuning or any additional training of the language model. 

\paragraph{Knowledge Updating.}
Despite the fact that LLMs like GPT-3 are pretrained on very large corpora, they are still far from perfect in terms of factual knowledge. On one hand, they still make factual mistakes even on domains that have seen before during pretraining (e.g., Wikipedia); on the other hand, they are pretrained on static corpora and hence their knowledge can become outdated. In order for LLMs to serve as reliable knowledge bases~\citep{Petroni2019LanguageMA} or power knowledge-intensive downstream applications~\citep{Petroni2021KILTAB}, it is important to keep LLMs' knowledge factually correct and up-to-update. A recent line of work attempts to edit factual knowledge in LLMs by making targeted modifications of the model's neurons~\citep{DeCao2021EditingFK,Mitchell2021FastME,Mitchell2022MemoryBasedME}. However, these methods are hard to be applied on GPT-3 since it is much larger in size and often treated as a black box without access to internal parameters. To address this issue, in this paper we explore the feasibility of performing in-context knowledge updating by directly appending relevant knowledge pieces in the prompt to guide model predictions. Since it has been shown that larger models are better at memorization~\citep{Carlini2022QuantifyingMA}, we analyze whether it is possible to make larger models forget their memorized knowledge and adapt to the new information presented in the prompt, especially when these two are in conflict. 
The idea of adding retrieved passages is conceptually similar to the line of work on retrieval-augmented methods for knowledge-intensive NLP~\citep{Lewis2020RetrievalAugmentedGF,Izacard2022FewshotLW,Guu2020REALMRL}. However, these methods still require supervised training while we focus on the setting of few-shot prompting with all the language model's parameters being frozen.

\section{Additional Results: Generalizability}
\label{appendix:generalization_baselines}

We provide full experimental results and comparisons with more baselines. 

\paragraph{MRQA}
Table~\ref{tab:mrqa_iid_full} and Table~\ref{tab:mrqa_ood_full} present detailed results on MRQA.  For baselines, we include results from the top-performing systems of the MRQA competition: D-Net~\citep{Li2019DNETAP} and Delphi~\citep{Longpre2019AnEO}, a recent adapter-based robust tuning method MADE~\citep{Friedman2021SingledatasetEF} as well as their multi-dataset finetuning baseline. We also report the finetuning and prompt tuning~\cite{Lester2021ThePO} result of using T5, which achieves state-of-the-art OOD transfer results on MRQA. Note that this T5 baseline only uses SQuAD as the in-domain training data. 
% Importantly, all baselines are not directly comparable with GPT-3 since they are trained by supervised finetuning on smaller-scale language models, while GPT-3 is much larger in model scale but requires no finetuning. The purpose of comparing with these baselines is to help readers understand how GPT-3 compares to other conventional models in terms of OOD generalization. 

\begin{table*}[t]
\small
\begin{center}
\setlength{\tabcolsep}{2mm}{
\begin{tabular}{ l c c c c c c c }
\toprule
   
    & SQuAD & HotpotQA & TriviaQA & NewsQA & SearchQA & NQ & Average \\
 \toprule
 D-Net & -- & -- & --& -- & -- & -- &  84.1 \\
 Delphi & -- & -- & --& -- & -- & -- &  82.3 \\
 MultiFT & 91.8 & 81.0 & 80.1 & 72.3 & 84.7 & 79.5 & 81.6 \\
 MADE & 91.9 & 80.7 & 80.1 & 71.8 & 84.5 & 79.5 & 81.4 \\
 T5-Finetune & 94.9 & -- & -- & --& -- & -- & -- \\
 T5-PromptTune & 94.8 & -- & -- & --& -- & -- & -- \\
 GPT-3 Source-P & 87.8 & 78.9 & 88.6 & 60.1 & 87.3 & 76.2 & 79.8 \\
 \bottomrule
\end{tabular}}
 \caption{Results on MRQA in-domain datasets. We use the F1 metric for all datasets. D-Net and Delphi results are taken from the MRQA system report~\citep{Fisch2019MRQA2S} which only reported the average performance. MultiFT and MADE results are taken from \cite{Friedman2021SingledatasetEF}. T5-Finetune and T5-PromptTune results are from the prompt tuning paper~\citep{Lester2021ThePO} which only trained on the SQuAD datsaet. GPT-3 few-shot performance slightly lags behind these supervised baselines in the in-domain setting.}
 \label{tab:mrqa_iid_full}
\end{center}
\end{table*}

\begin{table*}[t]
\small
\begin{center}
\setlength{\tabcolsep}{2mm}{
\begin{tabular}{ l c c c c c c c }
\toprule
   
    & BioASQ & DROP & DuoRC & RACE & RE & TextbookQA & Average \\
 \toprule
 D-Net & -- & -- & --& -- & -- & -- &  69.7 \\
 Delphi & -- & -- & --& -- & -- & -- &  68.5 \\
 MultiFT & 64.1 & 51.5 & 63.0 & 47.6 & 87.3 & 59.0 & 62.1 \\
 MADE & 66.5 & 50.9 & 67.2 & 47.8 & 86.7 & 58.5 & 62.9 \\
 T5-Finetune & 77.9 & \textbf{68.9} & 68.9 & 59.8 & 88.4 & 54.3 & 69.7  \\
 T5-PromptTune & 79.1 & 67.1 & 67.7 & 60.7 & 88.8 & 66.8 & 71.7 \\
 GPT-3 Source-P & \textbf{86.2} & 67.7 & \textbf{70.5} & \textbf{69.0} & 89.3 & \textbf{84.8} & \underline{\textbf{77.2}} \\
  GPT-3 Target-P & 85.9 & \textbf{68.9} & 69.7 & 65.4 & \textbf{91.0} & 82.1 & \underline{\textbf{77.2}} \\
 \bottomrule
\end{tabular}}
 \caption{Results on MRQA OOD datasets. ID-P (second last row) uses a fixed set of sampled demo examples from the in-domain datasets and evaluates on these OOD datasets, while OOD-P (last row) uses sampled demo examples from each of these OOD datasets for evaluation. GPT-3 significantly outperforms all other supervised baselines on these OOD datasets, moreover, using the in-domain prompt successfully transfers to OOD test data, achieving the same average performance as using demos examples drawn from these OOD datasets as the prompt.}
 \label{tab:mrqa_ood_full}
\end{center}
\end{table*}

\begin{table*}[t]
\small
\begin{center}
\setlength{\tabcolsep}{3.5mm}{
\begin{tabular}{ l c c c c c c }
\toprule
    & SST-2 & MNLI & RTE & QNLI & QQP & Average \\
    \midrule
\multicolumn{7}{c}{\textit{Clean Test Sets}} \\
\midrule
RoBERTa & 96.0 & 89.8 & 86.6 & 94.1 & 92.0 &  91.7 \\
ALBERT & 95.2 & 89.6 & 88.4 & 95.3 & 92.3 & 92.2 \\
DeBERTa & 96.3 & 90.9 & 90.2 & 94.9 & 92.3 & 92.9 \\
GPT-3 & 96.1 & 78.1 & 83.4 & 79.8 & 83.5 & 84.2 \\
\midrule
\multicolumn{7}{c}{\textit{Adversarial Test Sets}} \\
\midrule
RoBERTa & 58.5 & 45.2 & 45.4 & 52.5 & 57.1 & 51.7 \\
ALBERT & 66.8 & 48.0 & 73.0 & \textbf{63.8} & 56.4 & 61.6 \\
DeBERTa & 57.9 & 55.5 & 78.9 & 57.9 & 60.4 & 62.1 \\
GPT-3 & \textbf{78.4} & \textbf{56.4} & \textbf{87.7} & 57.4 & \textbf{66.7} & \textbf{69.3} \\
 \bottomrule
\end{tabular}}
 \caption{Results on the clean and adversarial test sets of AdvGLUE, we use accuracy as the metric for all datasets. For MNLI, we report the average performance on the matched and mismatched dev sets. The supervised baselines are trained on the clean training data, and GPT-3 uses few-shot prompts sampled from the clean training data. While GPT-3 few-shot prompting lags behind supervised models on clean test sets, it significantly outperforms the supervised models on the adversarial sets. That being said, we still note a performance drop of GPT-3 on the adversarial test sets when using clean demo examples.}
 \label{tab:advglue_full}
 % \vspace{-3mm}
\end{center}
\end{table*}

\begin{table*}
\small
\begin{center}
\setlength{\tabcolsep}{3.5mm}{
\begin{tabular}{ l c c c }
\toprule
& IMDB - Original & IMDB - Contrast & Gap$_{\downarrow}$ \\
\midrule
BERT & 93.8 & 84.2 & 9.6 \\
GPT-3 & 94.1 & 93.6 & 0.5 \\
\toprule
& QuoREF - Original & QuoREF - Contrast & Gap$_{\downarrow}$ \\
\midrule
XLNet & 70.5 & 55.4  & 15.1 \\
GPT-3 & 86.1 & 77.0 & 9.1 \\
\toprule
& BoolQ - Original & BoolQ - Contrast & Gap$_{\downarrow}$ \\
\midrule
RoBERTa & 86.1 & 71.1 & 15.0 \\
GPT-3 & 85.5 & 80.0 & 5.5 \\
 \bottomrule
\end{tabular}}
 \caption{Results on Contrast Sets. For IMDB and BoolQ, we report accuracy; for QuoREF, we report F1. Apart from the performance on the original and contrast sets of the three datasets, we also note the gap between performance on the original and contrast sets. We see a clear trend that GPT-3 incurs a smaller gap than the supervised models.}
 \label{tab:contrast_sets_full}
\end{center}
% \vspace{-3mm}
\end{table*}

\begin{table*}[t]
\small
\begin{center}
\setlength{\tabcolsep}{3.5mm}{
\begin{tabular}{ l c c c  }
\toprule
    & QQP (ID) & PAWS (OOD) & Gap  \\
    \midrule
BERT (supervised) & 91.3 & 40.1 & 51.2 \\
RoBERTa (supervised) & 89.0 & 39.5 & 49.5 \\
\midrule
Code-Davinci-002 (4-shots) & 78.2 & 80.5 &  -2.3 \\
Code-Davinci-002 (16-shots) & 83.5 & 73.7 & 9.8 \\
% Code-Davinci-002 (32-shots) & \\
\midrule
Text-Davinci-001 (16-shots) & 72.4 & 42.4 
 & 30.0  \\
Text-Curie-001 (16-shots) & 40.1 & 32.1 & 8.0 \\
 \bottomrule
\end{tabular}}
 \caption{Ablation for the impact of the number of demos and  different GPT-3 model variants on MNLI-HANS.}
 \label{tab:generalization_ablation}
 % \vspace{-3mm}
\end{center}
\end{table*}

\paragraph{AdvGLUE and Contrast Sets} 
Table~\ref{tab:advglue_full} and Table~\ref{tab:contrast_sets_full} present detailed results on AdvGLUE and Contrast Sets, where GPT-3 shows better generalization than supervised baselines.

% \begin{table}[t]
% \small
% \begin{center}
% \setlength{\tabcolsep}{4mm}{
% \begin{tabular}{ l c c c }
% \toprule
%     & BERT & RoBERTa & GPT-3  \\
%     \midrule
% \multicolumn{4}{c}{\textit{MNLI} $\rightarrow$ \textit{HANS}} \\
% \midrule
% MNLI & 86.2 & 89.1 & 77.6 \\
% HANS & 71.4 & 77.1 & 75.3 \\
% Gap$_{\downarrow}$ & 14.8 & 12.0 & 2.3 \\
% \midrule
% \multicolumn{4}{c}{\textit{QQP} $\rightarrow$ \textit{PAWS}} \\
% \midrule
% QQP & 91.3 & 89.0 & 83.5 \\
% PAWS & 40.1 & 39.5 & 73.7 \\
% Gap$_{\downarrow}$ & 51.2 & 49.5 & 9.8 \\
%  \bottomrule
% \end{tabular}}
%  \caption{Results on the challenge test sets from HANS and PAWS where spurious features from training data do not always hold. We find that GPT-3 few-shot prompting (demo examples in the prompt are randomly sampled from MNLI and QQP) incurs much smaller drops on the HANS and PAWS challenge sets as compared to supervised models trained on MNLI and QQP.}
%  \label{tab:spurious_full}
% \end{center}
% \end{table}

\begin{table}[t]
\small
\begin{center}
\setlength{\tabcolsep}{4mm}{
\begin{tabular}{ l c}
\toprule
  HANS Category  &  GPT-3 Acc. \\
\midrule 
Lexical Overlap - Entailment & 87.9  \\
Lexical Overlap - Non-Entailment & 96.9  \\
Subsequence - Entailment & 84.0 \\
Subsequence - Non-Entailment & 53.7 \\
Constituent - Entailment & 87.2  \\
Constituent - Non-Entailment & 42.2  \\
 \bottomrule
\end{tabular}}
 \caption{Breakdown of GPT-3 results on HANS by categories. All demo examples in the prompt are from MNLI. GPT-3 still suffers significant performance gaps between the entailment (bias-supporting) and non-entailment (bias-countering) subsets for the subsequence and constituent features in HANS.}
 \label{tab:hans_breakdown_full}
\end{center}
\end{table}

\paragraph{Spurious Correlation}
 Table~\ref{tab:hans_breakdown_full} shows full the performance breakdown on HANS based on the three spurious features. We can see that for the subsequence and constituent features in HANS, GPT-3 still suffers significant performance gaps between the bias-supporting and bias-countering subsets. This leaves curious questions like why such gaps only occur for certain bias features but not others, and how such spurious biases arise (most likely due to pretraining), and we leave a more thorough analysis of these questions to future work. 

In Table~\ref{tab:generalization_ablation}, we perform additional ablation on the impact of the number of demos and the different GPT-3 variants. With fewer demo examples from QQP, despite a slight drop on the QQP test set, GPT-3 actually remains robust (even higher accuracy on PAWS than the 16-shot results). On the other hand, using the Text-Davinci-001 (175B) and the smaller Text-Curie-001 (6.7B) performs far worse on both the QQP test set and the PAWS challenge test set.

\section{Additional Results: Social Biases}

We also break down the accuracy and bias scores of using different prompts in Table~\ref{tab:bbq_breakdown} by the different bias categories. We observe that there can be large differences across different categories. Moreover, we underlined the categories from which the demo examples come, and we observe that having same-category demos in the prompt does not correlate with the bias scores. For instance, we have bias-supporting examples from the Nationality category in the Ambig-Pro case but the bias score remains low, while the bias score for the Physical Appearance and Disability categories becomes much higher even when the biased examples are not from these categories.

\begin{table}[t]
\small
\begin{center}
\setlength{\tabcolsep}{2mm}{
\begin{tabular}{ l | c c | c c | c c}
\toprule
 & \multicolumn{2}{c}{Balanced}  & \multicolumn{2}{c}{Ambig-Pro} & \multicolumn{2}{c}{Ambig-Anti} \\
 & Acc$_{\uparrow}$ & Bias$_{\downarrow}$ & Acc$_{\uparrow}$ & Bias$_{\downarrow}$ & Acc$_{\uparrow}$ & Bias$_{\downarrow}$ \\
  \midrule
  % \multicolumn{7}{c}{\textit{Demo from Sexual Orientation and SES}} \\
  % \midrule
 SES & 96.4 / 74.2 & \underline{3.2 / 0.0} & 14.2 / 99.2 & \underline{-12.6 / 0.0} & 8.4 / 99.6 & \underline{-6.4 / 0.0} \\
Sexual orientation & 97.4 / 76.3 & \underline{1.6 / 1.2} & 1.9 / 98.4 & \underline{19.1 / -0.9} & 2.8 / 97.2 & \underline{17.2 / -0.9} \\
Religion & 97.0 / 75.8 & 1.8 / 0.5 & 0.4 / 98.0 & \underline{35.2 / 1.2} & 1.0 / 97.2 & \underline{24.6 / 1.6} \\
Race & 99.8 / 85.1 & -0.1 / -0.5 & 1.2 / 99.0 & 4.0 / 0.1 & 2.7 / 98.9 & 3.7 / 0.1 \\
Physical Appearance & 97.4 / 56.0 & 2.6 / 18.8 & 1.4 / 87.8 & 75.0 / 14.8 & 0.6 / 86.2 & 77.0 / 14.8 \\
Nationality & 98.2 / 80.8 & 1.4 / -11.6 & 1.0 / 99.0 & \underline{-0.2 / 0.0} & 1.0 / 98.6 & \underline{0.6 / 0.0} \\
Gender identity & 99.0 / 66.8 & 0.6 / -3.9 & 6.0 / 98.6 & 5.6 / 0.4 & 4.6 / 98.8 & 3.8 / 0.4 \\
Disability & 97.4 / 74.2 & 2.2 / 8.5 & 0.0 / 96.6 & 85.2 / 6.0 & 0.2 / 97.2 & 82.6 / 4.8 \\
Age & 82.2 / 76.6 & 13.0 / 8.1 & 0.0 / 95.4 & 52.0 / 12.4 & 0.4 / 95.6 & 48.4 / 12.0 \\
%    \midrule
%   \multicolumn{7}{c}{\textit{Demo from Age}} \\
%   \midrule
% SES & \\
% Sexual orientation & \\
% Religion & \\
% Race & \\
% Physical Appearance & \\
% Nationality & \\
% Gender identity & \\
% Disability & \\
% Age & \\
 \bottomrule
\end{tabular}}
 \caption{Breakdown of accuracy and bias score results on BBQ. For accuracy and bias scores, the first number represents the ambiguous set and the second number represents the disambiguated set. Underlined numbers indicate that there are demo examples in the prompt from the same bias category.} 
 \label{tab:bbq_breakdown}
\end{center}
\end{table}

\begin{table}[t]
\small
\begin{center}
\setlength{\tabcolsep}{2mm}{
\begin{tabular}{ l c c c c c }
\toprule
& DPR-BERT NQ & LM-Prob NQ & Self-Con NQ & LM-Prob TriviaQA & LM-Prob HotpotQA \\
\midrule
100\% & 36.1 & 40.5 & 40.2 & 73.8 & 29.8 \\
90\% & 38.0 & 43.7 & 44.3 & 78.3 & 32.7 \\
80\% & 39.5 & 46.8 & 48.7 & 81.7 & 36.0 \\
70\% & 40.6 & 50.2 & 53.1 & 84.1 & 39.7 \\
60\% & 41.2 & 53.7 & 57.8 & 86.5 & 43.5 \\
50\% & 41.9 & 58.8 & 62.0 & 88.5 & 47.6 \\
40\% & 43.3 & 63.3 & 66.0 & 90.5 & 52.1 \\
30\% & 46.1 & 70.2 & 71.2 & 92.5 & 56.5 \\
20\% & 49.2  & 77.4 & 74.7 & 93.7 & 61.6 \\
10\% & 60.1 & 83.1 & 77.0 & 95.4 & 68.1 \\
 \bottomrule
\end{tabular}}
 \caption{Selective prediction results. All numbers represent the accuracy (EM) at the corresponding coverage thresholds. For example, 100\% means performance on the entire test set while 10\% means the performance on the most confident 10\% predictions.} 
 \label{tab:selective_prediction_full}
\end{center}
\end{table}

\begin{table}[t]
\small
\begin{center}
\setlength{\tabcolsep}{3mm}{
\begin{tabular}{ l c c c  }
\toprule
& DPR-BERT NQ & LM-Prob NQ & Self-Con NQ \\
\midrule
100\% & 35.0 & 35.0 &  35.0 \\
90\% & 37.6 & 37.9 & 38.2  \\
80\% & 39.0 & 40.8 &  42.8 \\
70\% & 40.9 & 44.1 &  46.1 \\
60\% & 41.6 & 47.8 & 51.6 \\
50\% & 43.1 &  52.9 & 55.1 \\
40\% & 44.1 & 58.0 & 60.6 \\
30\% & 45.4 &  65.3 & 63.5 \\
20\% & 49.5 & 75.3 & 71.5 \\
10\% & 59.2 & 82.5 & 70.5 \\
 \bottomrule
\end{tabular}}
 \caption{Selective prediction results on a controlled set sampled from the original test set so that the accuracy on the sub-sampled test set is the same.} 
 \label{tab:selective_prediction_controlled}
\end{center}
\end{table}

\begin{figure*}[t]
\centering 
\begin{subfigure}{.49\linewidth}
  \centering
  \includegraphics[width=0.99\textwidth]{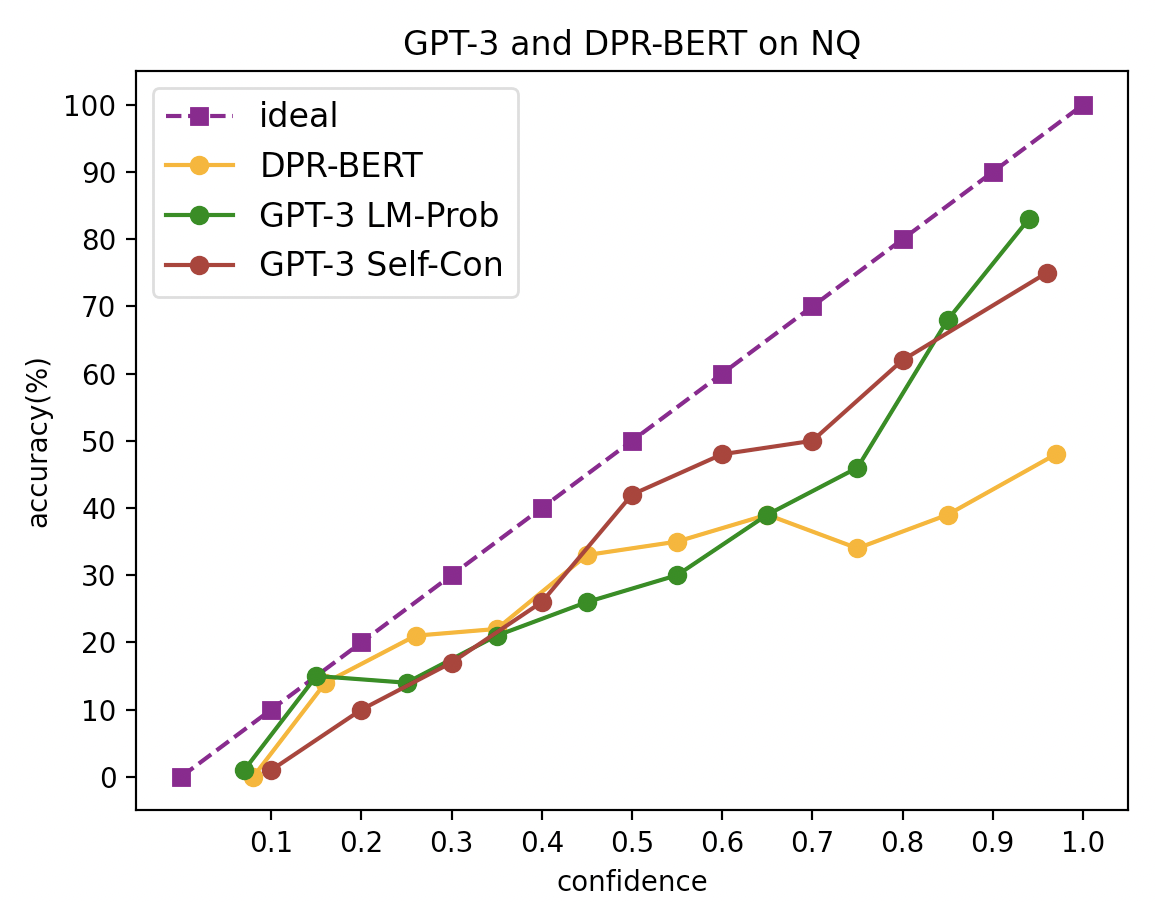}
%   \caption{Bias strength = 80\%.}
%   \label{fig:sub-first}
 \end{subfigure}
%  \hfill
  \begin{subfigure}{.49\linewidth}
  \centering
  \includegraphics[width=0.99\textwidth]{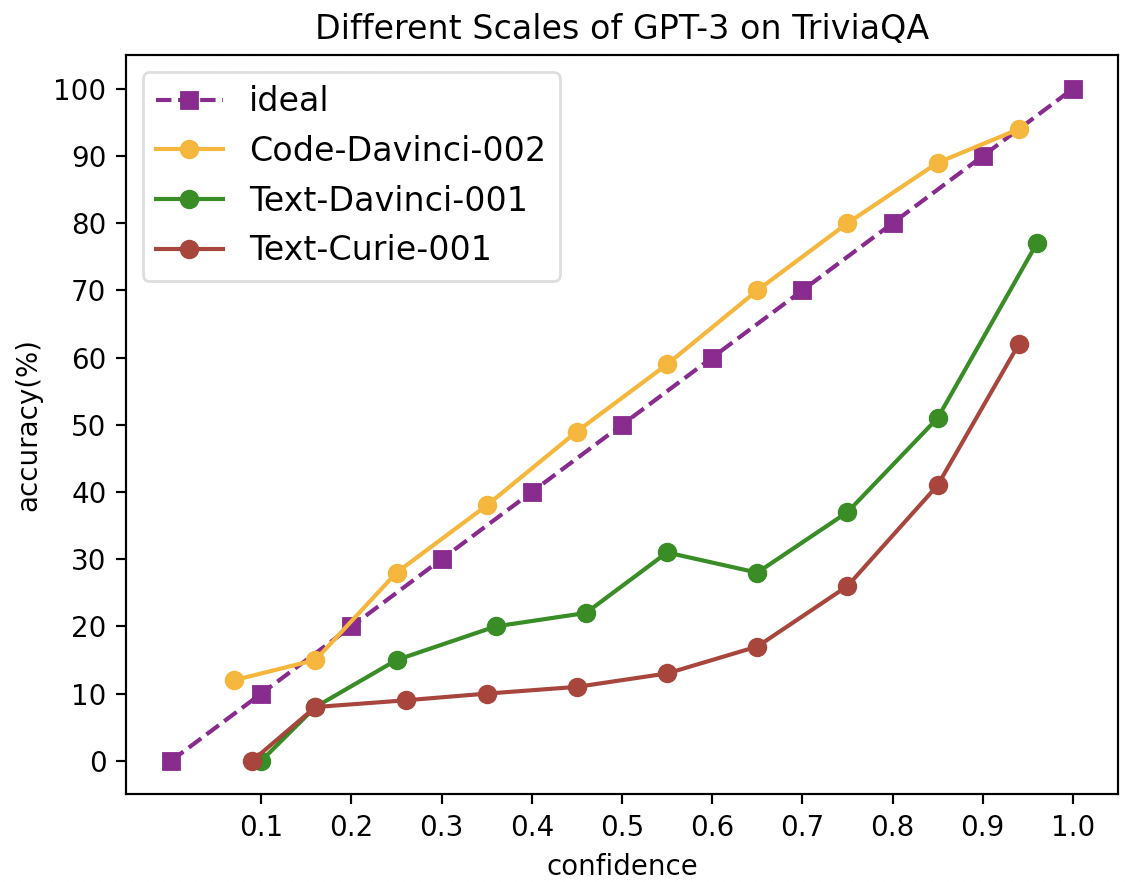}  
%   \caption{Bias strength = 90\%.}
%   \label{fig:sub-second}
\end{subfigure} 
\begin{subfigure}{.49\linewidth}
  \centering
  \includegraphics[width=0.99\textwidth]{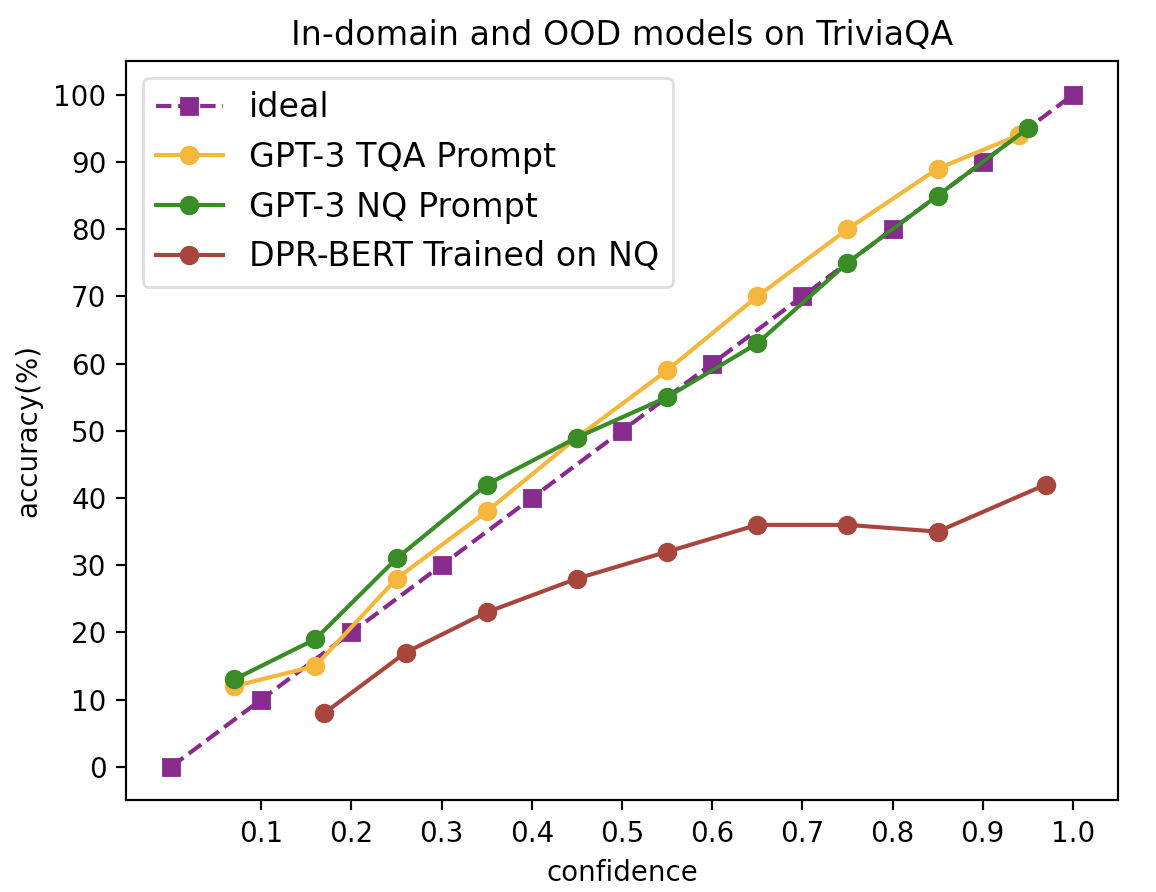}
%   \caption{Bias strength = 95\%.}
%   \label{fig:sub-first}
 \end{subfigure}
%  \hfill
  \begin{subfigure}{.49\linewidth}
  \centering
  \includegraphics[width=0.99\textwidth]{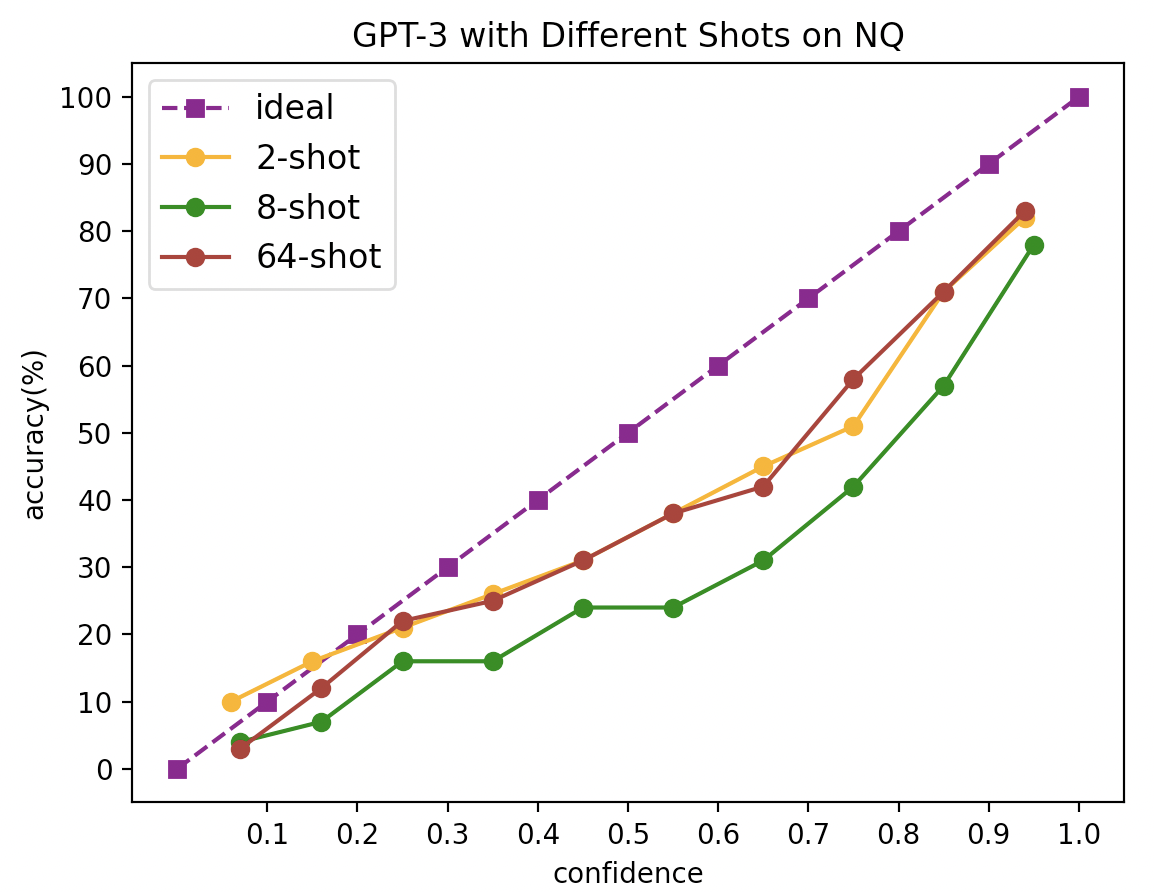}  
%   \caption{Bias strength = 100\%.}
%   \label{fig:sub-second}
\end{subfigure} 
\caption{Reliability diagrams for different calibration setups. x-axis: confidence of each bucket; y-axis: accuracy of each bucket.}
\label{fig:calibration_plots}
\end{figure*}

\section{Additional Results: Calibration}
\label{sec:more_calibration}

The full selective prediction results in Table~\ref{tab:selective_prediction_full} show that the confidence scores can be used to rank model predictions. We see a clear trend that the most confident predictions have much higher accuracy. 

The reliability diagrams in Figure~\ref{fig:calibration_plots} show that in most cases the calibration errors come from overconfidence where the predictions' confidence is higher than the  expected accuracy. It is also worth noting while OOD transfer is a big challenge for the calibration of supervised models where there tends to be overconfidence on the OOD test sets, GPT-3 exhibits similar calibration results when using in-domain or OOD demo examples as the prompt (bottom-left plot in Figure~\ref{fig:calibration_plots}). 

To further disentangle the impact of better accuracy and better calibration, we perform a controlled evaluation of selective prediction in Table~\ref{tab:selective_prediction_controlled} where we sub-sample the NQ test set so that the three calibration methods achieve the same accuracy on the test set. We see a clear trend that despite DPR-BERT and GPT-3 get same accuracy on this sub-sampled test set, DPR-BERT gets much higher accuracy on the most confident predictions indicating the usefulness of better calibration.

% \begin{table}[t]
% \small
% \begin{center}
% \setlength{\tabcolsep}{3.5mm}{
% \begin{tabular}{ l c  }
% \toprule
% & HotpotQA (EM / F1)  \\
% \midrule
%   \multicolumn{2}{c}{Closed-Book}\\
%   \midrule
% Standard Prompting & 18.0 / 28.1  \\
% CoT & 25.2 / 35.2  \\
% \midrule
%   \multicolumn{2}{c}{With Evidence Passages}\\
%   \midrule
% Standard Prompting & 71.7 / 85.3 \\
% CoT & 65.2 / 78.2 \\
%  \bottomrule
% \end{tabular}}
%  \caption{Results on HotpotQA with gold evidence passages appended. We see that when we provide the gold passages, CoT prompting brings no additional benefit (in this case it actually lowers the performance).}   
%  \label{tab:hotpot_gold}
% \end{center}
% \end{table}

\section{Additional Results: Knowledge Updating}

% \paragraph{The Benefit of Evidence Passages}
% The above analysis applies to the closed-book setting where we do not have the gold evidence passages. In our last experiment, we also experiment with the setting where we provide the gold evidence passages of HotpotQA questions in the prompt, in which case this problem is more like a reading comprehension setting. We also compare two prompting settings: standard prompting and CoT. From Table~\ref{tab:hotpot_gold}, we can see that adding the evidence passages achieves much higher QA performance than the closed-book model, as expected. Moreover, we find that when having gold passages in the prompt, CoT no longer brings benefits to multi-hop QA, and even causes performance drops. 

\begin{table}[t]
\small
\begin{center}
\setlength{\tabcolsep}{3.5mm}{
\begin{tabular}{ l | c c c | c }
\toprule
& Retain & Update & Other & Memorization Ratio \\
\midrule
  \multicolumn{5}{c}{\textit{NQ} with \textit{Code-Davinci-002}}\\
\midrule
T5 (supervised) & 20\% & 33\% & 47\% & 30\%  \\
GPT-3 Prompt $\langle$\textit{Q, A}$\rangle$ & 60.8\% & 25.8\% & 13.4\% & 70.2\% \\
GPT-3 Prompt $\langle$\textit{P, Q, A}$\rangle$ & 59.1\% & 25.4\% & 15.5\% & 70.0\% \\
GPT-3 Prompt $\langle$\textit{Q, A'}$\rangle$ & 10.4\% & 56.6\% & 32.9\% & 15.6\% \\
GPT-3 Prompt $\langle$\textit{P', Q, A'}$\rangle$ & 4.5\% & 85.4\% & 10.2\% & 5.0\% \\
\midrule
\multicolumn{5}{c}{\textit{SQuAD} with \textit{Code-Davinci-002}} \\
\midrule
GPT-3 Prompt $\langle$\textit{Q, A}$\rangle$ & 58.0\% & 29.1\% & 12.9\% & 66.6\% \\
GPT-3 Prompt $\langle$\textit{P, Q, A}$\rangle$ & 32.8\% & 52.9\% & 14.3\% & 38.2\% \\
GPT-3 Prompt $\langle$\textit{Q, A'}$\rangle$ & 15.4\% & 48.1\% & 36.5\% & 24.3\% \\
GPT-3 Prompt $\langle$\textit{P', Q, A'}$\rangle$ & 7.1\% & 84.8\% & 8.1\% & 7.8\% \\
\midrule
\multicolumn{5}{c}{\textit{NQ} with different versions of GPT-3} \\
\midrule
% \textit{Code-Davinci-002} & 4.5\% & 85.4\% & 10.2\% & 5.0\% \\
\textit{Text-Davinci-001} (175B) & 7.2\% & 57.9\% & 34.9\% & 11.0\% \\
\textit{Text-Curie-001} (6.7B) & 14.8\% & 40.0\% & 45.2\% & 26.9\% \\
 \bottomrule
\end{tabular}}
 \caption{In-context knowledge updating results for memorized answers in NQ and SQuAD. When giving counter-factual demo examples in the prompt, GPT-3 can update its answers around 85\% of the time with low memorization ratio (as compared to supervised models). Moreover, we find that larger models are better at in-context knowledge updating.}   
 \label{tab:memorization_full}
\end{center}
\end{table}

\begin{table}[t]
\small
\begin{center}
\setlength{\tabcolsep}{3mm}{
\begin{tabular}{ l c c }
\toprule
& Success Rate (Relevant Paraphrases) & Acc. Drawdown (Irrelevant Questions) \\
\midrule
  \multicolumn{3}{c}{Prompt: Original Examples Only}\\
\midrule
FEVER & 44.2 & 85.1 - 84.9 = 0.2 \\
zsRE QA & 92.9 & 40.0 - 39.7 = 0.3 \\
\midrule
  \multicolumn{3}{c}{Prompt: Original Examples + Edited Relevant Examples}\\
\midrule
FEVER & 100.0 & 83.9 - 48.6 = 35.3 \\
zsRE QA & 99.9 & 39.7 - 11.6 = 28.1 \\
\midrule
  \multicolumn{3}{c}{Prompt: Original Examples + Edited Relevant Examples + Edited Irrelevant Examples}\\
\midrule
FEVER & 99.9 & 84.0 - 83.5 = 0.5 \\
zsRE QA & 98.8 & 40.6 - 40.1 = 0.5 \\
 \bottomrule
\end{tabular}}
 \caption{Targeted in-context knowledge updating results for FEVER and zsRE.  From the last block of results, we see that when using a mixture of all three types of demo examples, GPT-3 is able to achieve very high knowledge edit success rate (99.9\% and 98.8\%) while incurring minimal drawdown (0.5\%) on irrelevant questions.}   
 \label{tab:knowledge_edit_full}
\end{center}
\end{table}

\subsection{Impact of Prompts for Memorization vs Updating}

For knowledge updating, we compare several different prompt designs as detailed below, for all cases, we randomly sample 16 demo examples as the prompt.
\begin{itemize}
    \item $\langle$\textit{Q, A}$\rangle$ : We use the original question-answer pairs in the prompt. 

    \item $\langle$\textit{P, Q, A}$\rangle$ : We use the original passage-question-answer triples in the prompt (i.e., the answer in the passage remains the original gold answer). 

    \item $\langle$\textit{Q, A'}$\rangle$ :  We use the question-answer pairs, but with the substitution entities as gold answers in the prompt. 
    
    \item $\langle$\textit{P', Q, A'}$\rangle$ : We use triples of the answer-substituted passage, the question, and the substitution answers in the prompt. 
\end{itemize}

As shown in Table~\ref{tab:memorization_full}, we find that the prompt design has a big impact on the knowledge updating behavior. In particular, showing only the original passage-question-answer triples ($\langle$\textit{P, Q, A}$\rangle$) still causes high memorization ratios, however, when prompting with counterfactual triples ($\langle$\textit{P', Q, A'}$\rangle$), GPT-3 can update 85\% of the time with much lower memorization ratios than a supervised model.

\subsection{Targeted In-Context Knowledge Updating}
The experiments in the previous section showed promise that GPT-3 can adapt to new knowledge given in the prompt when there is a conflict with its memorized knowledge. One missing aspect from the above analysis is whether we can perform \textbf{targeted} knowledge update: when given a piece of knowledge update, we expect the model to predict the updated answer for all questions related to that knowledge, but not change its answer for other unrelated questions. To assess model behavior on this front, we adopt an evaluation setup closer to recent knowledge updating literature~\citep{DeCao2021EditingFK,Mitchell2021FastME}.

\paragraph{Experiment Setup} We use two evaluation datasets from \cite{Mitchell2021FastME}: 1) We first use the fact checking dataset FEVER~\citep{Thorne2018FEVERAL}: each claim requires a binary true / false judgement. We create the edited label which is opposite to the originally predicted label from GPT-3. For example, for a test example, if the original GPT-3 prediction is true, then the new label for editing would be false. We present the knowledge update in the form of a natural sentence that supports the target label for editing. We then test whether GPT-3 predicts the target label for a \textbf{paraphrase} of the original test claim. We measure accuracy on these paraphrases as the editing \textbf{success rate}.  We sample a same-sized set from the training data that do not overlap with the test set as the set of unrelated questions. The intended behavior is that adding knowledge updates in the prompt does not impact performance on these unrelated questions. We measure performance drop on this unrelated set after and before adding knowledge updates as the accuracy \textbf{drawdown}. 2) We also use the zsRE question-answering dataset~\citep{Levy2017ZeroShotRE}. For each test question, the target label for editing is randomly sampled from predictions by a smaller QA model which is different from the original GPT-3 prediction. Similarly, we measure accuracy on paraphrased test questions as success rate, and accuracy drop on a set of randomly sampled non-overlapping training questions as the accuracy drawdown. 

\paragraph{Prompt Design}
 We compare several prompt designs (in particular what types of demo examples to use). For all cases, we sample 16 demos to use in the prompt. 

 \begin{itemize}
     \item Original Examples Only: We only sample the original QA pairs (without editing information). 

     \item Original + Edited Relevant Examples: We include demos examples for both original QA pairs as well as for questions with edited answers. 

     \item Original + Edited Relevant + Edited Irrelevant Examples: We include demo examples covering all possible cases: the original QA pairs, QA pairs with knowledge update and updated answer, as well as QA pairs with knowledge update but original answer (where the question is unrelated to the knowledge update). 
 \end{itemize}

\paragraph{Results}
From Table~\ref{tab:knowledge_edit_full}, we see that different prompts give vastly different results. Specifically, using only the original examples in the prompt leads to relatively poor success rate (especially on FEVER), while adding edited relevant examples in the prompt leads to better success rate, it leads the model to over-rely on the knowledge updates even on irrelevant questions. However, when incorporating all cases of original examples, edited relevant and irrelevant examples in the prompt, GPT-3 is able to achieve high editing success rate and low drawdown on irrelevant questions.

\end{document}